\documentclass[12pt,reqno,twoside]{amsart}
\usepackage{amsfonts,amsmath,amssymb}
\usepackage{mathrsfs,mathtools}
\usepackage{enumerate}
\usepackage{hyperref}
\usepackage{esint}
\usepackage{graphicx}
\usepackage{bm}
\usepackage{commath}
\usepackage{esint}
\usepackage{placeins}
\usepackage{flafter}
\usepackage{algorithm}
\usepackage{algpseudocode}
\usepackage{tikz}
\usepackage[textsize=small]{todonotes}
\usepackage[dvips,bottom=1.4in,right=1in,top=1in, left=1in]{geometry}
\usepackage[latin1]{inputenc}
\usepackage{comment}
\usepackage{subfigure}

\DeclareMathAlphabet{\mathpzc}{OT1}{pzc}{m}{it}
\numberwithin{equation}{section}

\makeatletter
\def\eqnarray{\stepcounter{equation}\let\@currentlabel=\theequation
\global\@eqnswtrue
\tabskip\@centering\let\\=\@eqncr
$$\halign to \displaywidth\bgroup\hfil\global\@eqcnt\z@
  $\displaystyle\tabskip\z@{##}$&\global\@eqcnt\@ne
  \hfil$\displaystyle{{}##{}}$\hfil
  &\global\@eqcnt\tw@ $\displaystyle{##}$\hfil
  \tabskip\@centering&\llap{##}\tabskip\z@\cr}
\def\endeqnarray{\@@eqncr\egroup
      \global\advance\c@equation\m@ne$$\global\@ignoretrue}

\numberwithin{equation}{section}

 \usepackage[textsize=small]{todonotes}
\title{Dynamic Reconstruction from Neuromorphic Data} 
\date{\today}

\thanks{
This work is partially supported by NSF grant DMS-2408877, the Air Force Office of Scientific Research (AFOSR) under Award NO: FA9550-22-1-0248 and the Office of Naval Research (ONR) under Award NO: N00014-24-1-2147. 
}

\author{Harbir Antil}
\address{H. Antil, D. Blauvelt, D. Sayre. The Center for Mathematics and Artificial Intelligence
(CMAI) and Department of Mathematical Sciences, George Mason University,
Fairfax, VA 22030, USA.}
\email{hantil@gmu.edu, dblauvelt@gmu.edu, dsayre@gmu.edu}
\author{Daniel Blauvelt}
\author{David Sayre}

\begin{document}				

\begin{abstract}	
Unlike traditional cameras which synchronously register pixel intensity, neuromorphic sensors only register `changes' at pixels where a change is occurring asynchronously. This enables neuromorphic sensors to sample at a micro-second level and efficiently capture the dynamics. Since, only sequences of asynchronous event changes are recorded rather than brightness intensities over time, many traditional image processing techniques cannot be directly applied.  Furthermore, existing approaches, including the ones recently introduced by the authors, use traditional images combined with neuromorphic event data to carry out reconstructions.  The aim of this work is introduce an optimization based approach to reconstruct images and dynamics only from the neuromoprhic event data without any additional knowledge of the events. Each pixel is modeled temporally. The experimental results on real data highlight the efficacy of the presented approach, paving the way for efficient and accurate processing of neuromorphic sensor data in real-world applications.
\end{abstract}


\maketitle

\section{Introduction} \label{s:Intro}

Event cameras, a category of neuromorphic sensors, utilize an asynchronous sampling mechanism to capture data based on changes in light intensity at each pixel. Specifically, an event is triggered at a pixel only when the change in light intensity surpasses a predefined threshold, resulting in no recorded events in the absence of scene changes \cite{gallego2020event,HAntil_DSayre_2023a}. In scenarios characterized by camera or object movement, the event camera captures intensity changes at each pixel with a temporal resolution on the order of microseconds. The output of a neuromorphic camera can be mathematically conceptualized as a spatio-temporal point process, with an event at a pixel represented as a binary-valued function of time \cite{gallego2018unifying,huang2023progressive}.
This paper delves into the challenges and opportunities of translating this event data into frame-based images. Unlike fixed-rate sampling in traditional cameras, neuromorphic camera sample at rates dependent on scene dynamics. This allows neuromorphic cameras to avoid under-sampling of swiftly changing scenes or redundant over-sampling of slowly changing ones.

Recently, a substantial body of research has been dedicated to the challenge of image reconstruction from event data, leading to the development of various methodologies spanning multiple domains. One approach involves deep learning techniques, which leverage the power of neural networks to interpret and reconstruct images from the sparse and asynchronous data provided by event cameras as shown by \cite{paredes2021back,scheerlinck2020fast,wang2019event,yang2023learning,zou2021learning}.
Another significant strand of research focuses on inverse problem approaches, which treat the reconstruction process as a mathematical problem solvable through optimization techniques. These methods, explored in works such as \cite{HAntil_DSayre_2023a,pan2019bringing,zhang2021formulating}, approach reconstruction by solving a set of equations that describe the relationship between the event data and the underlying image.

The method proposed in this paper reimagines the reconstruction of frame-based images from event data by treating it as an initial value problem within the framework of ordinary differential equations (ODEs). In this approach, the temporal evolution of luminosity at each pixel is modeled as an ODE, with event data providing insights into the temporal derivative of luminosity rather than its absolute value. A significant challenge is the unknown initial condition, the true luminosity of each pixel at the beginning of the observation period. While previous methods, such as those in \cite{HAntil_DSayre_2023a, pan2019bringing}, rely on standard camera images to estimate these initial conditions, and \cite{zhang2021formulating} employs spatial correlations by warping event images, our method takes a different approach. We avoid spatial correlations altogether and instead formulate a system of equations that captures the temporal evolution of each pixel's luminosity. By applying a least squares approach, we solve this system, enabling the reconstruction of scene dynamics without any prior knowledge of the initial luminosity.

The structure of the paper is as follows: 
Section~\ref{s:not} introduces basic notations and discusses the previous work. 
Section \ref{s:opt} presents the proposed model, outlines the reconstruction algorithm, and provides an alternate derivation that establishes that our optimization problem indeed has a physical interpretation. Section \ref{s:numres} details the parameters and procedures we employed during our computational experiments and explains the role and physical significance of the regularization parameter as well as ways it can be computed. Section \ref{s:numres} also provides three real-world dataset examples for which we apply the proposed method demonstrating its efficacy. It concludes by briefly identifying existing challenges with the proposed method and opportunities for future work.

\section{Notation and Preliminaries}
\label{s:not}
\subsection{Neuromorphic Cameras}
\subsubsection{Overview}
Neuromorphic cameras consist of individual pixels that independently monitor changes in light intensity in real-time. The sampling of light intensity occurs at a rate of microseconds ($\mathcal{O}(\mu s)$), and events are recorded if the detected light intensity, with respect to the intensity at the same pixel at the previous time, surpasses a predefined hardware threshold $c$ \cite{HAntil_DSayre_2023a}. We denote the light intensity at pixel $(x,y)$ and time $s$ as $q_{xy}(s)$. The high sampling rate and the independent nature of the camera's pixels make neuromorphic cameras resilient to exposure issues and image blurring.

\subsubsection{Data}\label{s:Data}

Event cameras output data in the form $(x,y,s,p)$. Here $(x, y)$ represent a pixel location at which an event has occurred, with a total of $N_x \times N_y$ points (or camera resolution). Furthermore,  $s \in \{s_1,s_2, \dots, s_n \, : \, s_1 < s_2 < \dots < s_n \} \subset \mathcal{I}$ represents the distinct times at which events occurred within the interval $\mathcal{I}$ and $n$ is the total number of events over all pixels. To properly define the polarity $p$ we must first consider a subset of the events for a specific pixel $(\bar x, \bar y)$. Recall that not every pixel will generate events at the same time. Thus for a specific pixel $(\bar x,\bar y)$ we can describe the set of times at which events occurred only for the specific pixel $(\bar x, \bar y)$ as $S_{(\bar x \bar y)} = \{s_j : j =1, \dots, n \text{ and }(\bar x, \bar y, s_j, p) \text{ is a recorded event} \}$. Now we can define the polarity $p$ as follows:
\begin{equation}\label{eq:polarity}
{p}_{\bar x \bar y}(s_j) := \begin{cases}
+1, & \log \left( \frac{{q}_{\bar x \bar y}(s_j)}{{q}_{\bar x \bar y}(s_{j-1})}\right) \geq c\\[.2in]
-1, & \log \left( \frac{{q}_{\bar x \bar y}(s_j)}{q_{\bar x \bar y}(s_{j-1})}\right) \leq -c.
\end{cases}
\end{equation}
We note that $q_{\bar x \bar y}(s_0)$ represents the initial luminosity at camera start-up. We also assert $q_{\bar x \bar y}(s_j) \in \mathbb{R}^+\setminus\{0\} $ for any $j$. Here we do not include $0$ since that would represent the camera operating in absolute darkness. Recall that the value $c$ represents the camera threshold, and no event is recorded if ${p}_{x y}(\cdot) \in (-c, c)$. For the remainder of the paper we will omit the subscript $(x,y)$ representing the pixel location unless necessary for clarity.

\subsection{Existing Model, Algorithm, and Limitations} \label{s:existing}
In \cite{HAntil_DSayre_2023a}, our approach combined event data from neuromorphic sensors with standard camera images to reconstruct scenes, aiming to leverage the temporal sensitivity of event data alongside the spatial detail of conventional images. However, this method's effectiveness hinged significantly on the quality of the standard camera images, which posed challenges in synchronization and could compromise overall reconstruction accuracy. Moreover, implementing this approach involved solving a complex bi-level optimization problem, adding further intricacy to the computational process.

The model presented in this paper represents a significant departure from previous approaches by simplifying the reconstruction process exclusively through event data, thereby obviating the requirement for standard camera images. Unlike conventional methods that rely on both event data and traditional images to reconstruct scenes, this model directly uses the temporal sequence of event timestamps. It captures variations in pixel luminosity by aggregating events across arbitrarily defined time intervals, enabling the reconstruction of pixel intensities at any specific timestamp independently of camera imagery. Furthermore, to our knowledge, no other studies investigating event frame reconstruction have approached the problem solely from a temporal perspective.

A potential limitation of the proposed model includes the sensitivity to errors in event data. Since the reconstruction relies solely on the temporal sequence of event timestamps, inaccuracies or missing events could impact the fidelity of the reconstructed image. Therefore, ensuring the accuracy and completeness of event data becomes crucial for achieving reliable results. Additionally, another limitation is the requirement for a sufficient amount of temporal data. Adequate coverage of events over time intervals is essential to accurately capture the dynamics of pixel luminosity changes. Insufficient temporal data may lead to incomplete or inaccurate reconstructions.

\section{Proposed Model and Algorithm}\label{s:opt}

\subsection{Model\label{s:model}}
Consider an arbitrary pixel $( x, y)$ and the corresponding set of event time-stamps, $S_{ x y}$, as described in \ref{s:Data}. Next, denote $\phi_{s}(t)$ to be a unit bump function centered at $s$. 
Then we can define the continuous function 
\begin{equation}\label{eq:exy}
e_{xy}(t) := \sum_{j=1}^n {p_{xy}(s_j)} \cdot  \phi_{s_j}(t) \in \mathbb{R} .
\end{equation}
Next, we define the function $E_{xy}(t_i,t_k)$ as the accumulation of events between arbitrary times $t_i,t_k \in \mathcal{I}$ 
\begin{equation}\label{eq:Exy}
{E}_{xy}(t_i,t_k) = \int_{t_i}^{t_k}e_{xy}(t) \; dt .
\end{equation}
Recall that events are recorded when the luminosity at a given pixel changes above a threshold. Thus we can model the change in a pixel luminosity between arbitrary times $t_i$ and $t_k$ as:
\begin{equation}\label{eq:vxy}
{v}_{xy}(t_k) - {v}_{xy}(t_i) = {E}_{xy}(t_i,t_k) \, .
\end{equation}
Where $v(t_k)$ and $v(t_i)$ represent the luminosity of the pixel at time $t_k$ and $t_i$ respectively. Now, consider the sequence of discrete times $t_l$, $l =1,\dots,N_t$ that divide the interval $\mathcal{I}$ such that $0 \leq t_1 < t_2 <\cdots <t_{N_{t-1}}\leq t_{N_t}$. Here $N_t\in \mathbb{N}$ and represents the number of frames to generate. We can construct tensors $\bm v$, $\bm E \in \mathbb{R}^{N_x \times N_y \times N_t}$ as:

\[
\bm{v}_{xy} = \begin{pmatrix}
    v_{xy}(t_1)\\
    v_{xy}(t_2)\\
    \vdots\\
    v_{xy}(t_{N_t})
\end{pmatrix}  \quad \text{and} \quad
\bm{E}_{xy} = \begin{pmatrix}
    E_{xy}(t_1,t_2)\\
    E_{xy}(t_2,t_3)\\
    \vdots\\
    E_{xy}(t_{N_t-1},t_{N_t})
\end{pmatrix}  \, .
\]
Notice that, here $\bm{v}_{xy}$ and $\bm{E}_{xy}$ are vectors in
time for pixel $(x,y)$. 
This allows us to form a system of equations by: 
\begin{equation}
	\bm{A}\bm{v}_{xy} = \bm{E}_{xy}
\end{equation}
with $\bm A$ defined as:
\begin{equation}
\bm{A} = \begin{pmatrix}
\begin{array}{rrrrrrrr}
-1  &   1 &      &     &  & &  &  \\
&  -1 &  1  &     &  &  &  &  \\
&     &  & \ddots & \ddots &   &  &   \\
&     &     &    &   & -1 & 1 &   \\		
&     &     &    &   &    & -1 & 1 \\
\end{array}	
\end{pmatrix}.
\end{equation}
Subsequently, we arrive at the following optimization problem to solve 
for pixel intensity through time for each pixel $(x,y)$ 
\begin{equation}\label{eq:in}
\bm{v}_{xy}=\operatorname*{argmin}_{\bm v_{xy}} 
\frac{1}{2} \| {\bm A} \bm v_{xy}  -  \bm E_{xy}\|_2^2 
+ \frac{1}{2} \|\bm\lambda \cdot  \bm v_{xy}\|_2^2 .
\end{equation}
Here $\bm \lambda = \text{diag}(\lambda_1, \lambda_2, \dots \lambda_{N_t}) \in \mathbb{R}^{N_t \times N_t}$ with $\lambda_i > 0, \; i=1,\dots,N_t$.
Notice that for any $\bm \lambda$ \eqref{eq:in} is uniquely solvable by:
\begin{equation}
\bm{v}_{xy} 
= (\bm A^\top \bm A  + \bm \lambda^\top \bm \lambda)^{-1} \bm A^\top \bm E_{xy}
= \bm K^{-1} \bm{A^\top}\bm E_{xy}\label{eq:v}
\end{equation}
where $\bm K:= (\bm A^\top \bm A  + \bm \lambda^\top \bm \lambda)$.

\subsection{Alternate Derivation}
Recall from \ref{s:Data} that $S_{(\bar{x} \bar{y})}$ is the set of 
events that occur at pixel $(\bar{x}, \bar{y})$ with 
\begin{equation}\label{eq:pos}
\frac{{q}_{\bar x \bar y}(s_j)}{{q}_{\bar x \bar y}(s_{j-1})} > 0 ,
\end{equation}
for any $s_j \in S_{(\bar{x} \bar{y})}$. Notice that, \eqref{eq:pos} 
is due to our assumption ${q}_{\bar x \bar y}(s_j) \in \mathbb{R}^+$
for every $s_j \in S_{(\bar{x} \bar{y})}$.
For $\Delta_j \in \{-c,c\}$ and an error term $\epsilon_j$, from 
\eqref{eq:polarity} we can write: 
\begin{align}
&\log \left( \frac{{q}_{\bar x \bar y}(s_j)}{{q}_{\bar x \bar y}(s_{j-1})}\right) = \triangle_j + \epsilon_j, 
\label{eq:qxylog} \\
\implies \quad &\log ( {q}_{\bar x \bar y}(s_j)) - \log ( {q}_{\bar x \bar y}(s_{j-1})) = \triangle_j + \epsilon_j \, . \notag
\end{align}
Recall the definition of polarity from $\eqref{eq:polarity}$:
\begin{equation*}
{p}_{\bar x \bar y}(s_j) := \begin{cases}
+1, & \log \left( \frac{{q}_{\bar x \bar y}(s_j)}{{q}_{\bar x \bar y}(s_{j-1})}\right) \geq c\\[.2in]
-1, & \log \left( \frac{{q}_{\bar x \bar y}(s_j)}{q_{\bar x \bar y}(s_{j-1})}\right) \leq -c.
\end{cases}
\end{equation*}
Consider the following: let ${p}_{\bar x \bar y}(s_j) = +1 $ then 
$\log \left( \frac{{q}_{\bar x \bar y}(s_j)}{{q}_{\bar x \bar y}(s_{j-1})}\right) \geq c$ which using \eqref{eq:qxylog} implies that 
\begin{align}\label{eq:pxyDelta} 
{p}_{\bar x \bar y}(s_j) = 1 = \frac{\Delta_j}{c} \, ,
\end{align}
because $\Delta_j = c$ in this case. 
Same argument applies for ${p}_{\bar x \bar y}(s_j) = -1$. 

Now, if we consider arbitrary times $s_i, s_k$ with $k > i$, then we can expand to the following:
\begin{align}
&\log ( {q}_{\bar x \bar y}(s_k)) - \log ( {q}_{\bar x \bar y}(s_{i})) 
= \sum_{l =i}^k \left( \triangle_l + \epsilon_l \right) \notag \\
\implies &c^{-1} \Big(\log ( {q}_{\bar x \bar y}(s_k)) - \log ( {q}_{\bar x \bar y}(s_{i})) \Big) 
= c^{-1} \sum_{l =i}^k \left( \triangle_l + \epsilon_l \right) \notag \\
\implies &\log ( ({q}_{\bar x \bar y}(s_k))^{\frac{1}{c}}) - \log (( {q}_{\bar x \bar y}(s_{i}))^{\frac{1}{c}}) 
= \sum_{l =i}^k \left( \frac{\triangle_l}{c} + \frac{\epsilon_l}{c} \right).
\label{eq:approxQ}
\end{align}
Notice that from $\eqref{eq:exy}$ and $\eqref{eq:Exy}$ we have that 
\begin{align*}
{E}_{\bar x \bar y}(s_i,s_k) = \int_{s_i}^{s_k}\sum_{l=i}^k {p_{\bar x \bar y}(s_l)} \cdot  \phi_{s_l}(t) dt \, .
\end{align*} 
Substituting \eqref{eq:pxyDelta} above leads to
\begin{align}\label{eq:approxE}
{E}_{\bar x \bar y}(s_i,s_k) &= \int_{s_i}^{s_k}\sum_{l =i}^k 
\frac{\triangle_l}{c} \cdot  \phi_{s_l}(t) dt 
= \sum_{l =i}^k 
\frac{\triangle_l}{c} \int_{s_i}^{s_k} \phi_{s_l}(t) dt 
\approx \sum_{l =i}^k 
\left( \frac{\triangle_l}{c} + \frac{\epsilon_l}{c} \right).
\end{align}
Here the approximation comes from approximating the integral via a sum. 
Then using \eqref{eq:vxy}, \eqref{eq:approxQ} and \eqref{eq:approxE} we have: 
\begin{align}\label{eq:vconnectq}
{v}_{\bar x \bar y}(s_k) - 	{v}_{\bar x \bar y}(s_i) = {E}_{\bar x \bar y}(s_i,s_k) 
\approx \sum_{l =i}^k \left(\frac{\triangle_l}{c} + \frac{\epsilon_l}{c} \right)
= \log ( ({q}_{\bar x \bar y}(s_k))^{\frac{1}{c}}) - \log (( {q}_{\bar x \bar y}(s_{i}))^{\frac{1}{c}}). 
\end{align}
Recall that the solution to optimization problem \eqref{eq:v} is given by $\bm v_{\bar{x}\bar{y}}$. Equation~\eqref{eq:vconnectq} implies that the difference between two components (representing two different times $s_k$ and $s_i$) of $\bm v_{\bar{x}\bar{y}}$, in fact approximates the difference in the same time components of the true luminosity $\bm q_{\bar{x}\bar{y}}$.

\section{Numerical Results}\label{s:numres}

\subsection{Experimental Setup \label{s:eset}}

This section focuses on how the dataset was prepared during our experiments.
\begin{description}\itemsep5pt 
    \item[Data Cube] The tensor $\bm{E}$ defined above contains the neuromorphic images that come from an event camera. 
    Depending on the event camera used, the entries of $\bm{E}$ may take on various ranges of values to include $[0,255]$ or $[-255,255]$. Since $\bm{E}$ is composed of a sequence of images over time, we build a three-dimensional array $\bm{dc}$ following the procedure described in \cite{HAntil_DSayre_2023a}. That is, for each pixel, $(x,y)$, in $\bm{E}$ we have
        \begin{equation*}
            \bm{dc}_{x,y}^{\pi} = \sum_{\omega = 1 + (\pi-1)\cdot k}^{\pi \cdot k} p_{x,y}^{\omega}
        \end{equation*}
    where $\omega$ are discrete times at which events occur and $\pi = 1,...,r$. Since event data is captured on a nearly continual basis we construct $\bm{dc} \in \mathbb{R}^{N_x \times N_y \times r}$ to obtain a more practically sized data representation. Our approach is to effectively compress the data in time by choosing $r,k \in \mathbb{N}$ such that $\nu \approx r \cdot k$ where we take
        \begin{equation*} 
        r = \left \lfloor \frac{\nu}{k} \right \rfloor .
        \end{equation*} 
    We generally chose $r$ and $k$ to obtain a value for $\nu$, the total number of events used in our reconstructions, that produced the highest quality reconstructions, but such that experiment computations executed in minutes on a standard commercial personal computer. We generally used $k=150$ in our numerical experiments.
    
    \item[Regularization Parameter $\bm \lambda$]
    Once the data cube, $\bm{dc}$, is built, we reconstruct the latent images by solving the least squares problem \eqref{eq:in}. Notice that $\eqref{eq:in}$ has two unknowns, $\bm v$ and $\bm \lambda$. In this section we formulate $\bm \lambda$ as a function of the data (events). 
    \subsubsection{Physical meaning of $\bm \lambda$}
    We make note that $\bm \lambda$ determines the amount of weight given to the data, $\bm{dc}$, at 
    each time slice, for each pixel. 
  If $\bm \lambda$ is set too low for pixels that have relatively more registered changes in luminosity by the event camera over a given time period, of high dynamics, then the reconstructed image can appear very blurry and washed out. In contrast, if $\bm \lambda$ is set too large for pixels that have relatively fewer registered changes in intensity by the event camera over a given time, of low dynamics, then the reconstruction may have a large error for those pixels. 

    \subsubsection{Proposed ways to calculate $\bm \lambda$.}
    As stated above, care must be taken to choose $\bm \lambda$ appropriately in order to generate accurate reconstructions. Below we provide various options. 
    We note that we must impose the restriction $\lambda_i \neq 0$ for any $i$.\\
    \begin{enumerate}
    
    \item Sigmoid function: 
        \begin{equation*}
            \lambda_i(\bm{dc}^i_{xy}) = \frac{1}{1 + \exp\big\{-|\bm{dc}^i_{xy}|\big\}}
        \end{equation*}
     \item Max (in absolute value): 
        \begin{equation*}
            \lambda_i(\bm{dc}^i_{xy}) = \max\{\epsilon, |\bm{dc}^i_{xy}|\} \quad \text{where } \epsilon > 0.
        \end{equation*}
    \item Exponential of absolute value
        \begin{equation*}
            \lambda_i(\bm{dc}^i_{xy}) = \exp\big\{|\bm{dc}^i_{xy}|\big\}.
        \end{equation*}
\end{enumerate}

    \item[Example 1: Office] This experiment used a dataset in which the neuromorphic camera is handheld by a person in an office setting with a goal to subject the sensor to rotational motions \cite{gallego2017accurate}. Notice that the person moves around the scene. 
    Here we use $\nu \approx 2.5M$ events and the sigmoid function, option (a) above, to compute $\bm{\lambda}$. As can be seen in the sequence depicted in Figure \ref{f:office}, our method recovers high levels of detail where distinct objects along with finer-grain details,  depth, perspective, and lighting are immediately evident. Notice that the facial features of the person are distinguishable as well as the different objects on the desk. Viewing clockwise from the computer in the upper right of the desk, a food canister, a couple of pens, a small stack of books, a quadcopter and an open book are distinguishable.

\begin{figure}[!h]
    \centering
    \includegraphics[width=.325\textwidth]{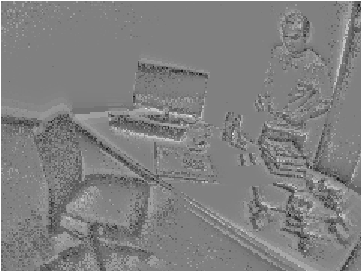}
    \includegraphics[width=.325\textwidth]{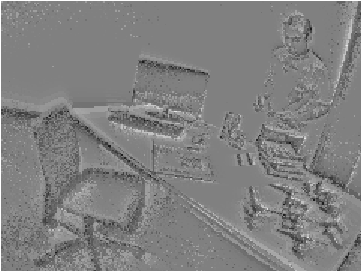}
    \includegraphics[width=.325\textwidth]{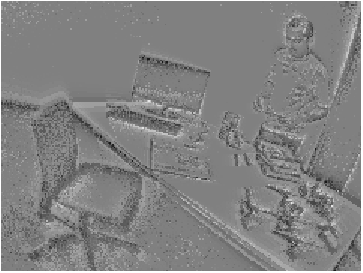}
    \includegraphics[width=.325\textwidth]{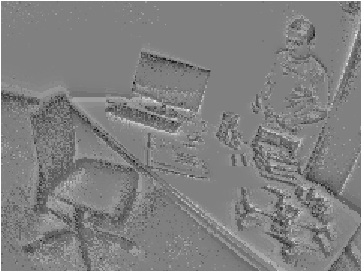}
    \includegraphics[width=.325\textwidth]{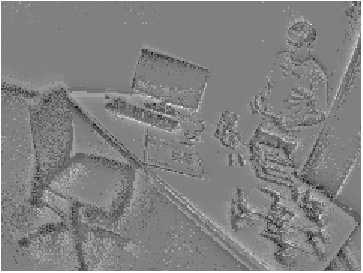}
    \includegraphics[width=.325\textwidth]{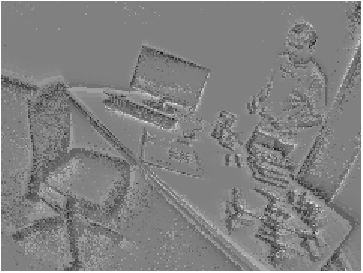}
    \includegraphics[width=.325\textwidth]{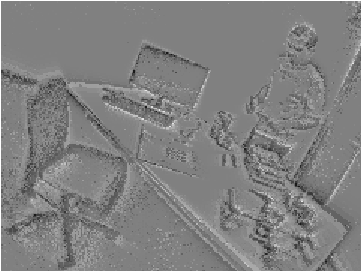}
    \includegraphics[width=.325\textwidth]{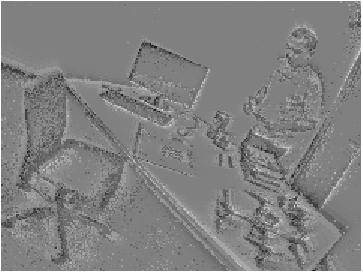}
    \includegraphics[width=.325\textwidth]{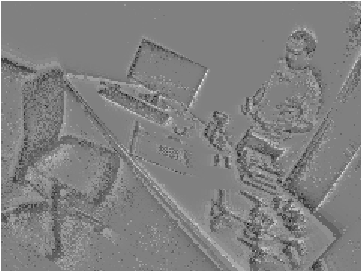}
    \includegraphics[width=.325\textwidth]{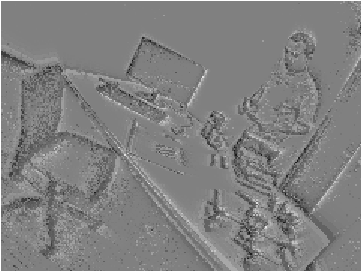}
    \includegraphics[width=.325\textwidth]{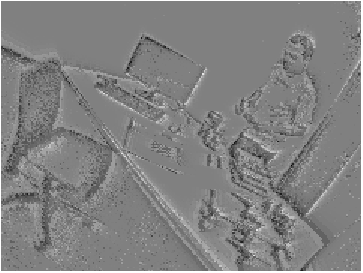}
    \includegraphics[width=.325\textwidth]{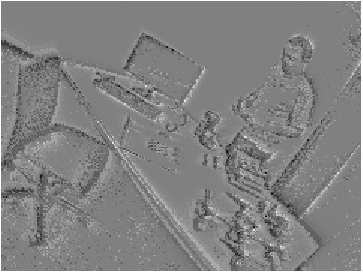}
    
    \caption{Dynamic sequence of reconstructed images showing a person moving in front of desk while event camera is rotating about scene. Throughout the sequence, the proposed method recovers good quality for the overall scene and also recovers depth and even object-level details.}
    \label{f:office}
\end{figure}
    
    \item[Example 2: Cup] For our second experiment, we applied the proposed approach to a dataset which captures the ballistics of a bullet hitting a cup on a table  \cite{rebecq2021highspeed}. Our image reconstructions for the cup data used $\nu \approx 350k$ events. Figure \ref{f:cup} depicts a close-up event sequence of a standard-sized ceramic coffee cup being shot while sitting on table that extends past the edges of the frame in a nondescript room. Reading from left-to-right, top-to-bottom, our method captured the initial exit spray of the bullet at the end of the first row. By the middle of the second row, a major fracture near the bullet's exit point is visible by a distinct white spot surrounded by the spreading ceramic spray. In the subsequent images, finer details of the fracture spreading are recovered into the seeming explosion and disintegration of the cup's visible form. Figure \ref{f:cup_cm} is the same sequence applied with a false color map to highlight the recovered dynamic details. Notice that by the end of the sequence, table edges and the room corner are much less visible which likely results from the combination of the neuromorphic camera only registering changes in intensity and our method for the regularization parameter which suppresses windows over pixels with low dynamics. For this reconstruction we formulated $\bm \lambda$ using the maximum in absolute value approach, option (b) above. In addition, the images were post processed using the \verb|mat2gray| function in Matlab to visually enhance the dynamic range.

\begin{figure}[!h]
    \centering
    \includegraphics[width=.325\textwidth]{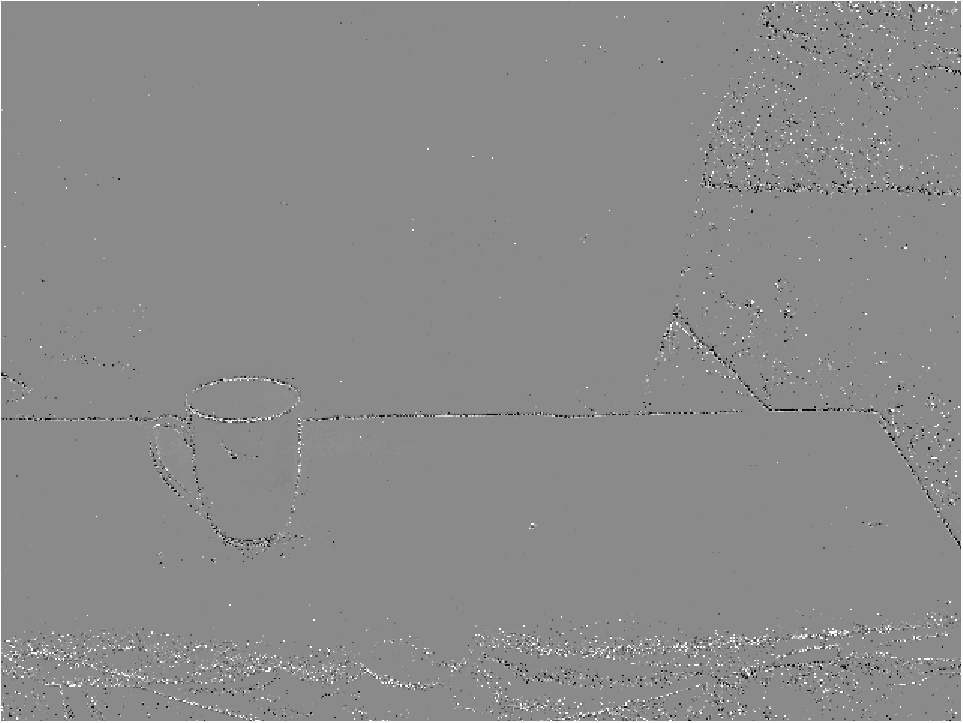}
    \includegraphics[width=.325\textwidth]{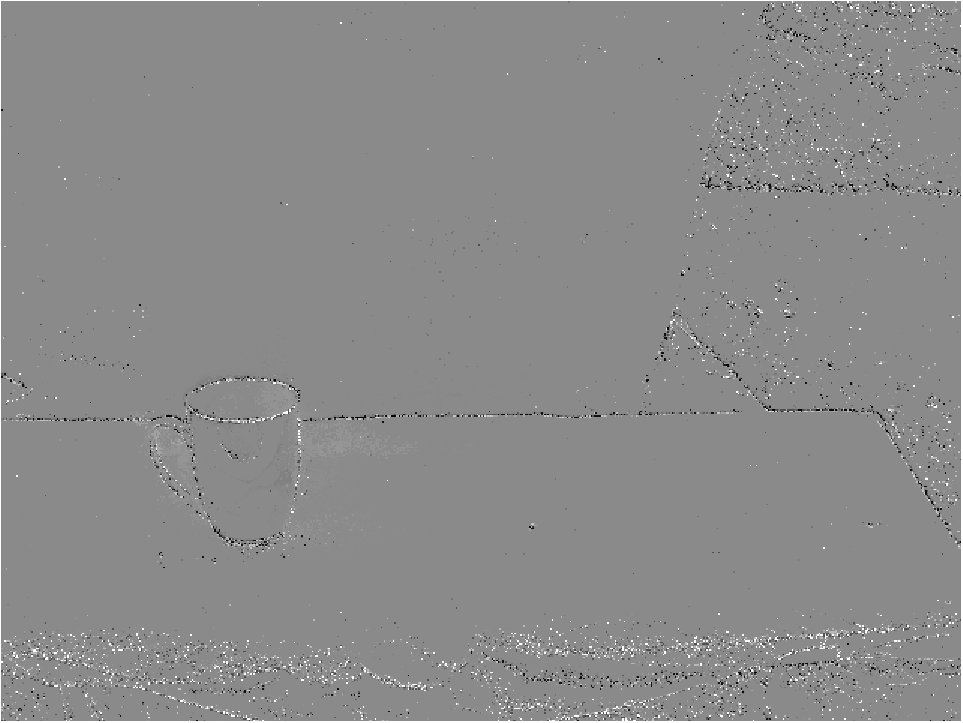}
    \includegraphics[width=.325\textwidth]{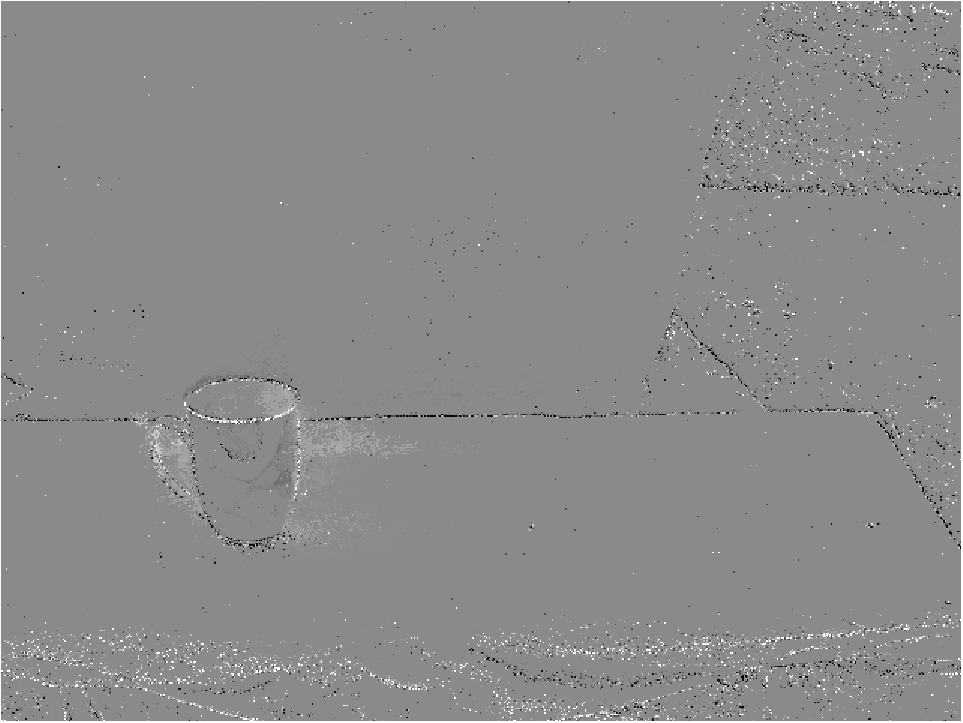}
    \includegraphics[width=.325\textwidth]{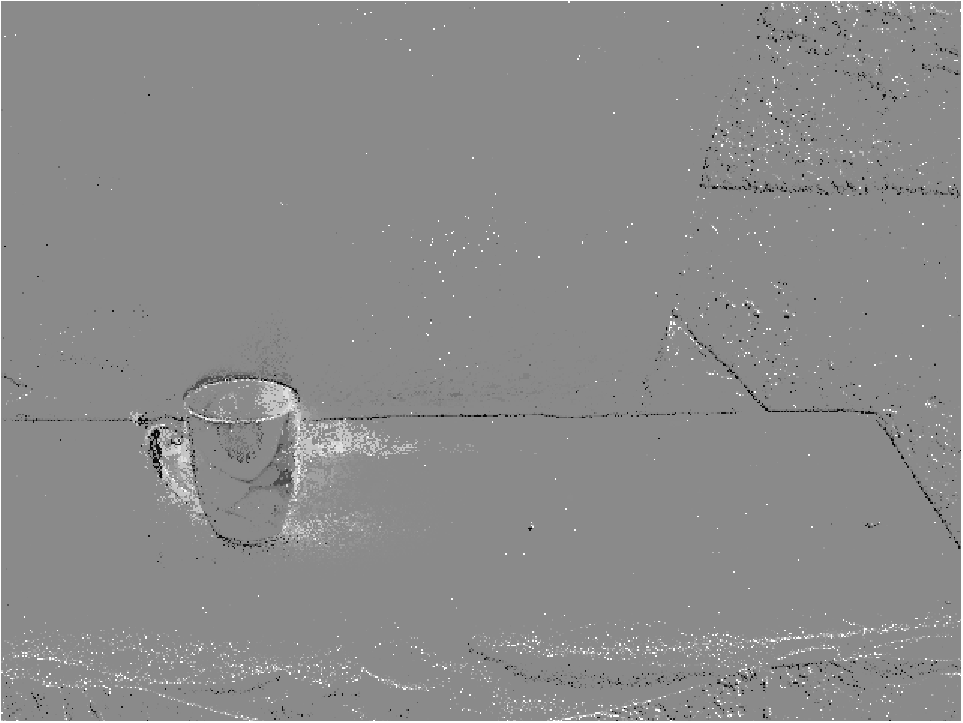}
    \includegraphics[width=.325\textwidth]{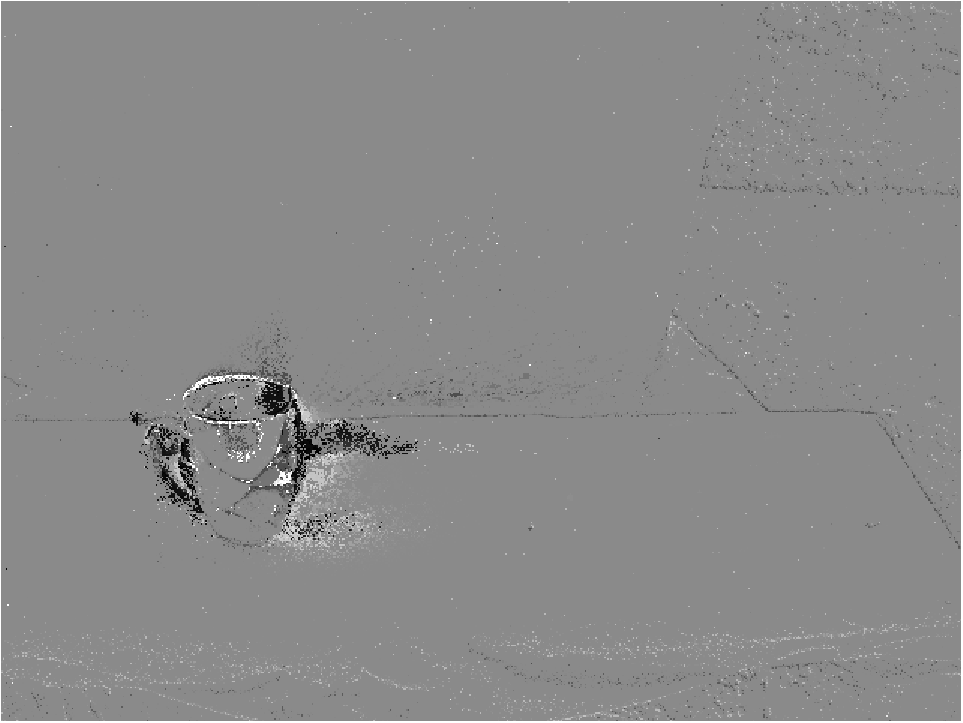}    
    \includegraphics[width=.325\textwidth]{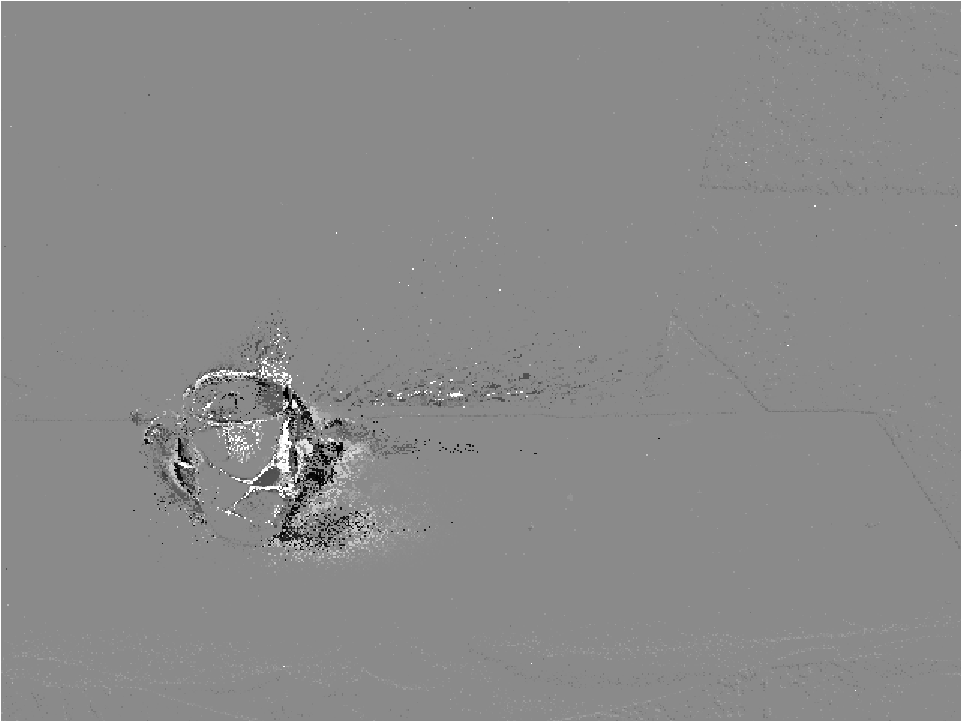}
    \includegraphics[width=.325\textwidth]{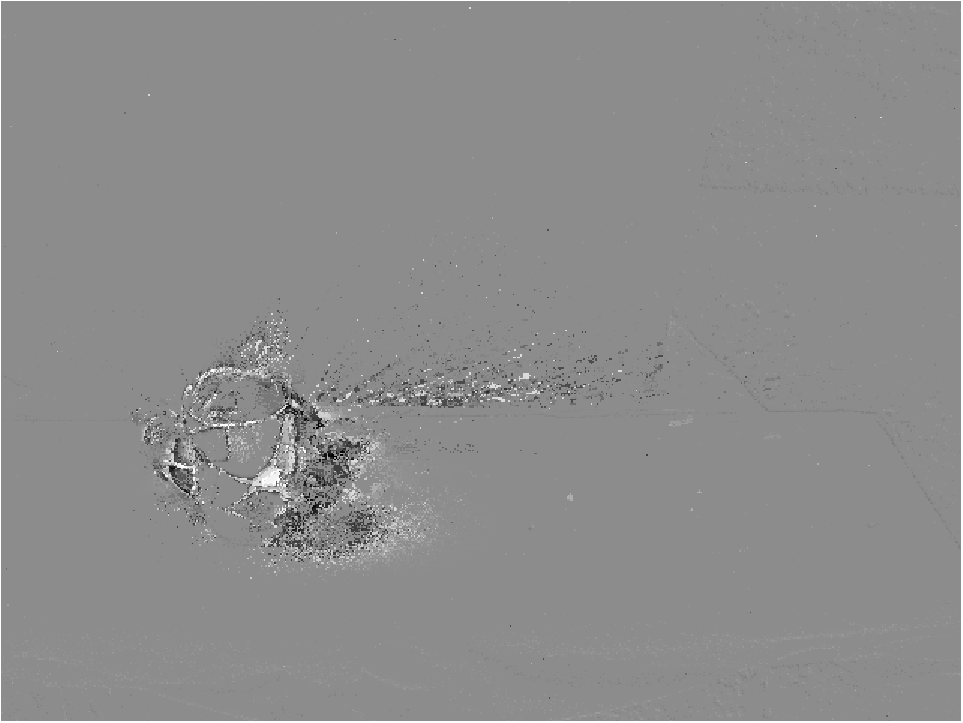}
    \includegraphics[width=.325\textwidth]{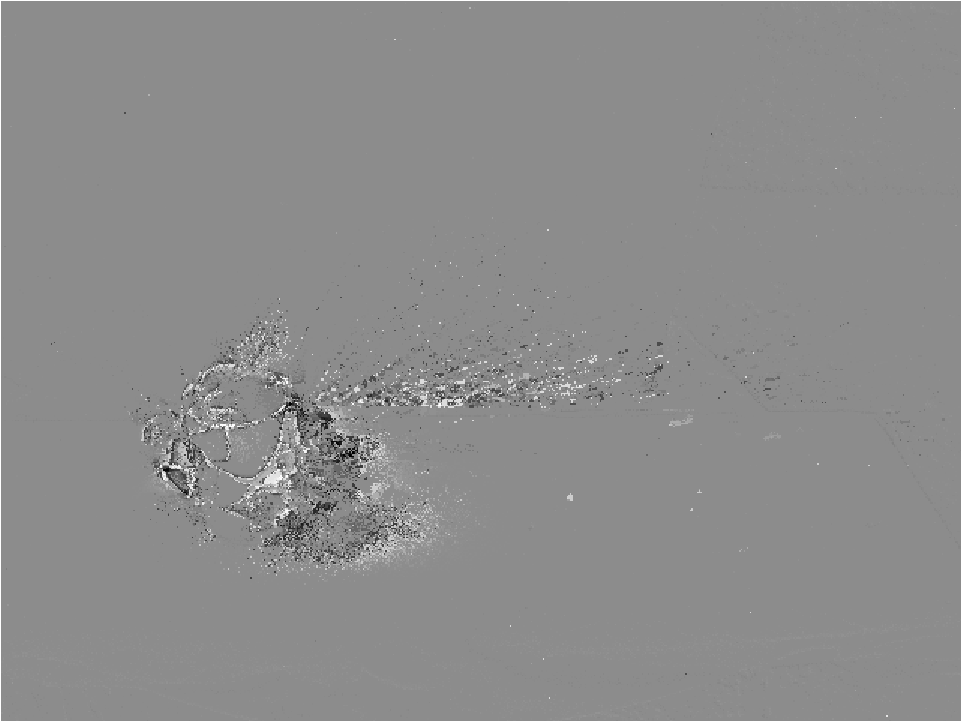}
    \includegraphics[width=.325\textwidth]{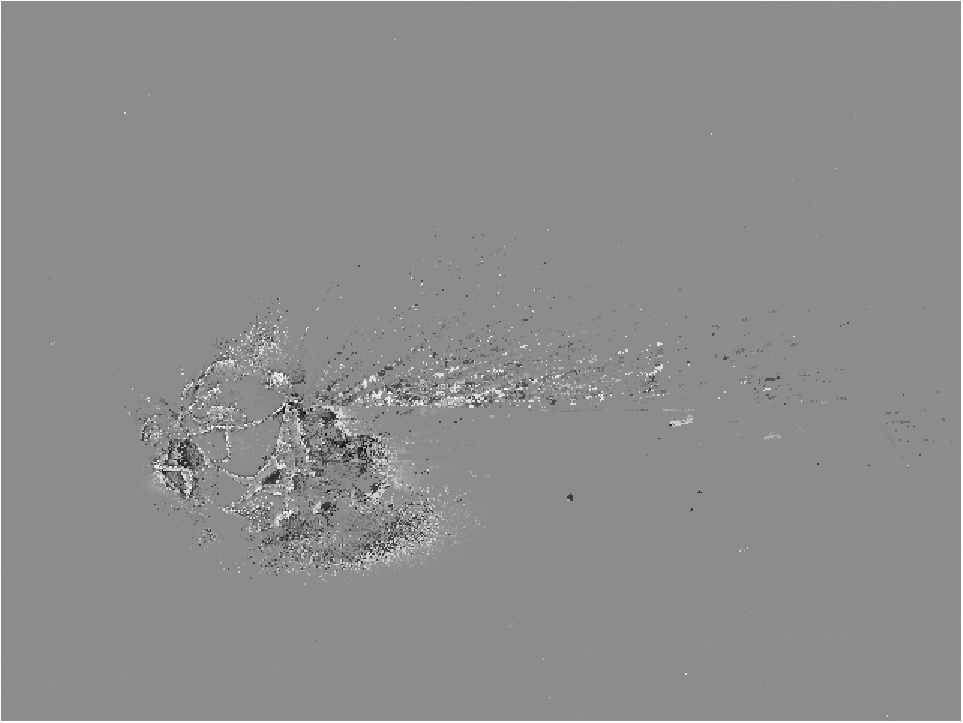}
    \includegraphics[width=.325\textwidth]{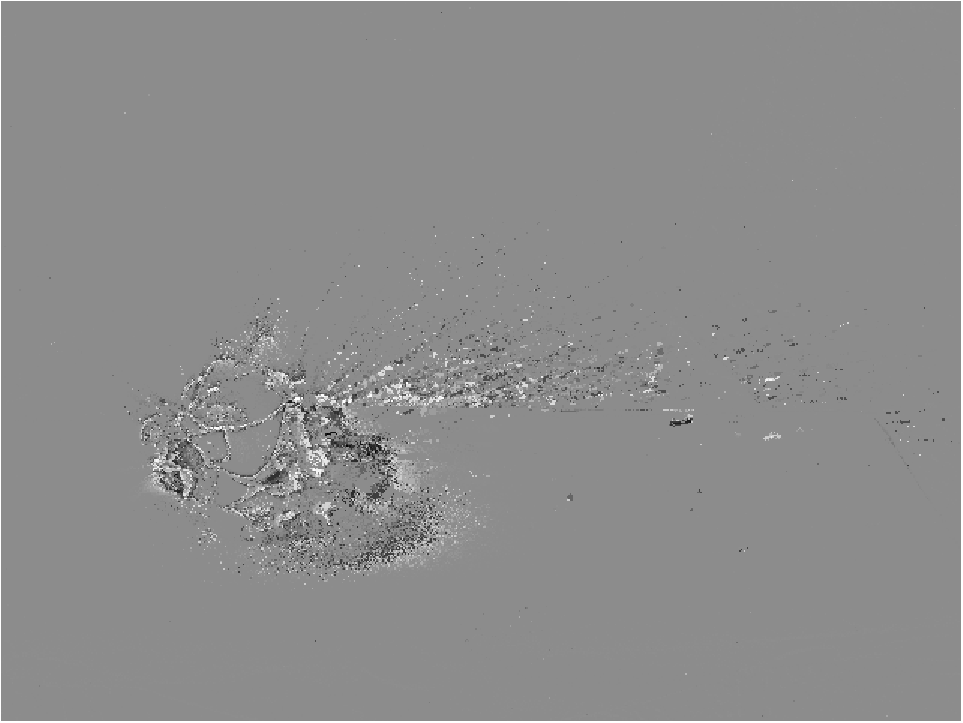}
    \includegraphics[width=.325\textwidth]{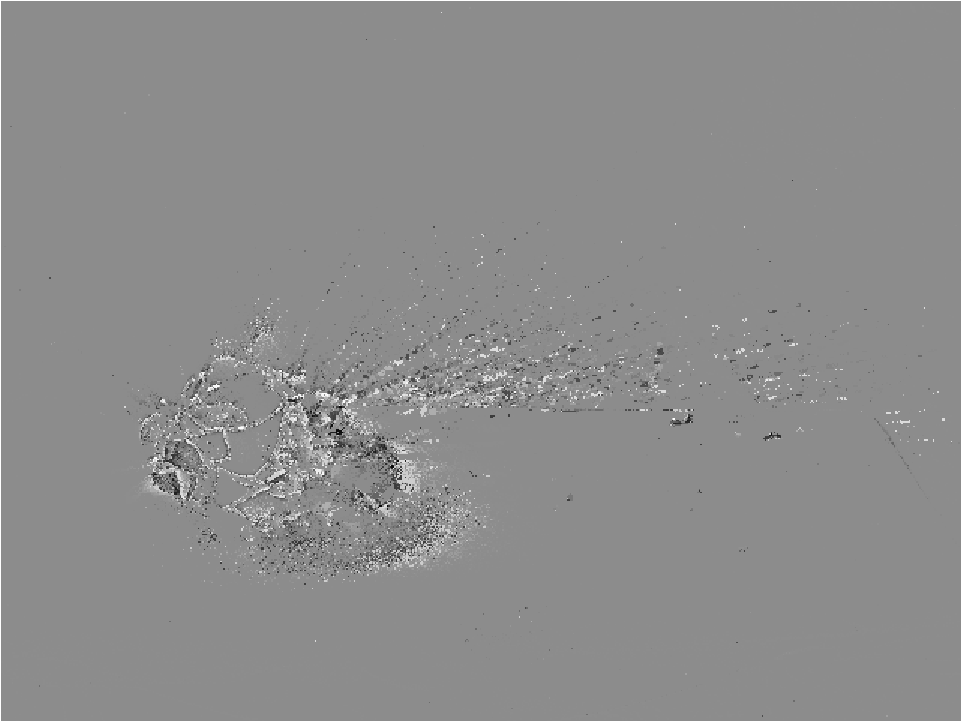}   
    \includegraphics[width=.325\textwidth]{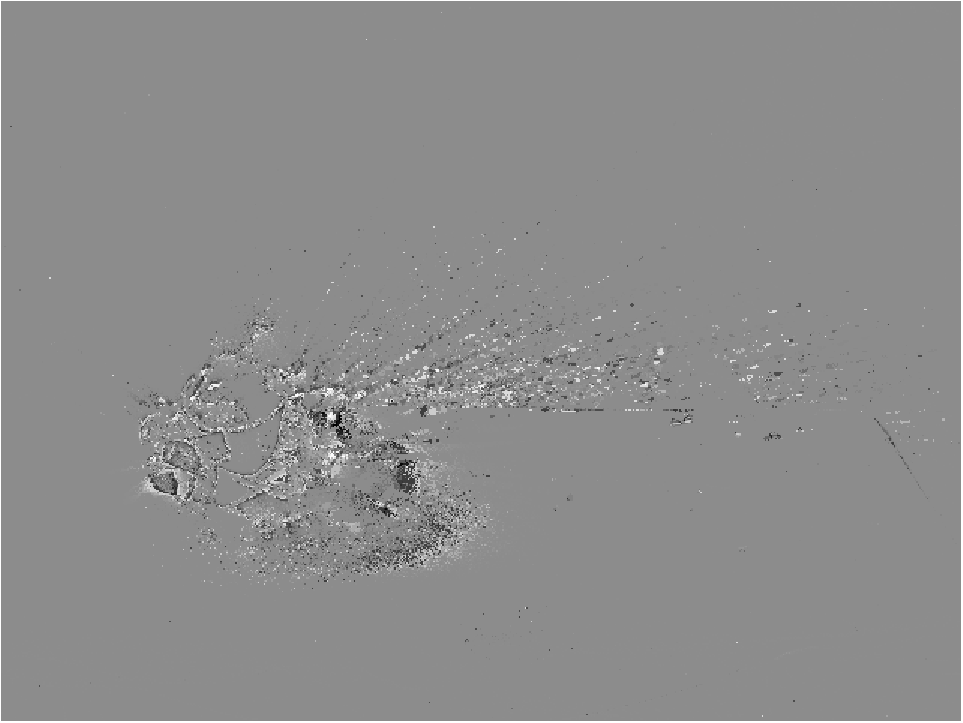}
    \includegraphics[width=.325\textwidth]{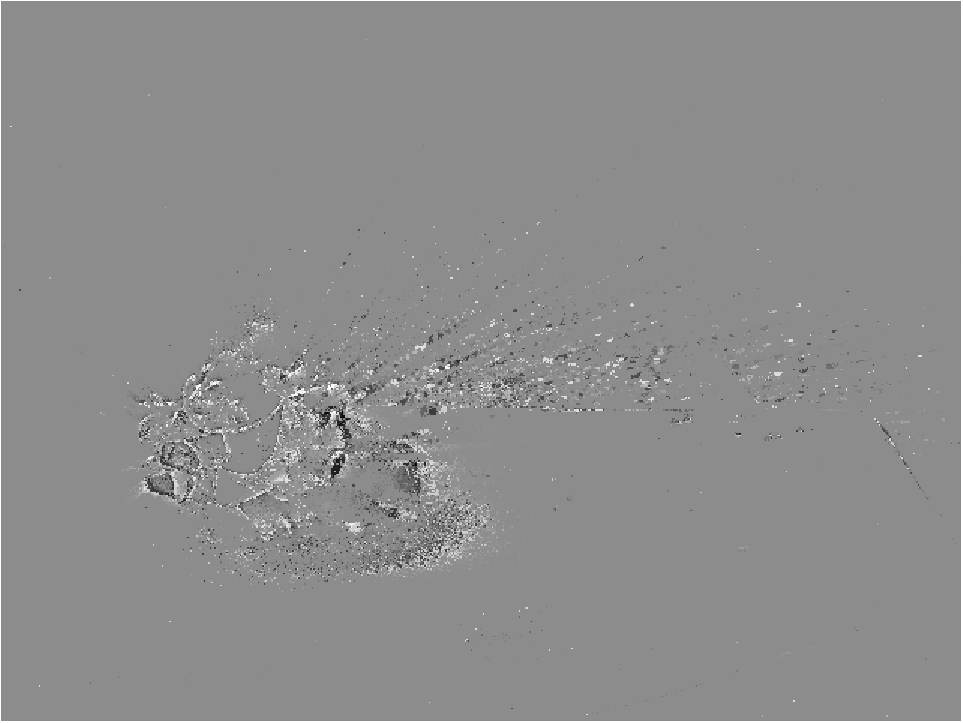}
    \includegraphics[width=.325\textwidth]{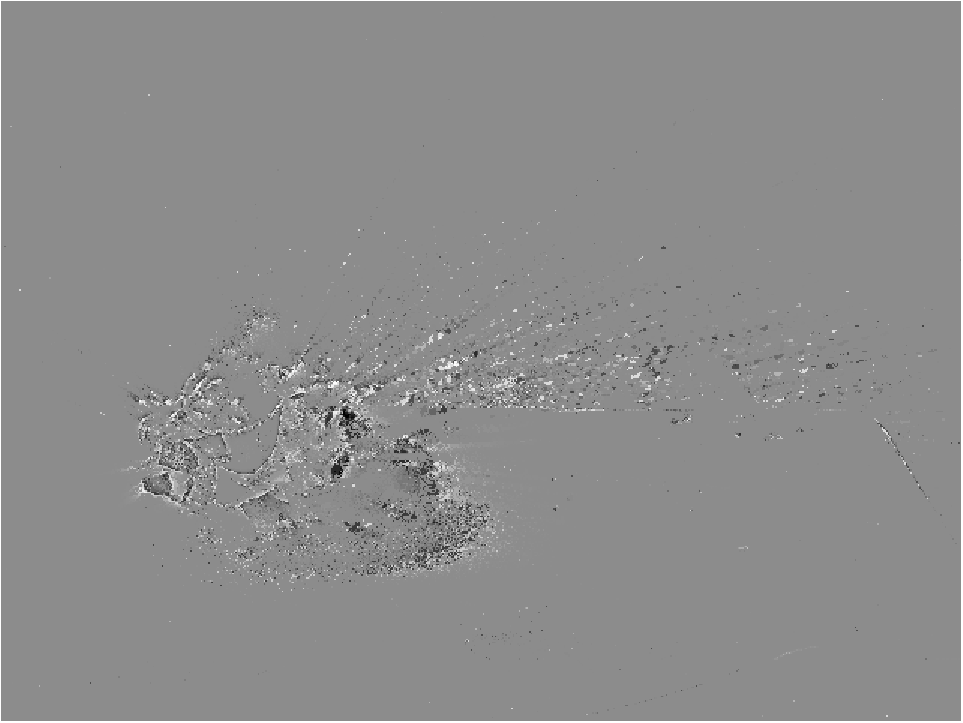}   
    \includegraphics[width=.325\textwidth]{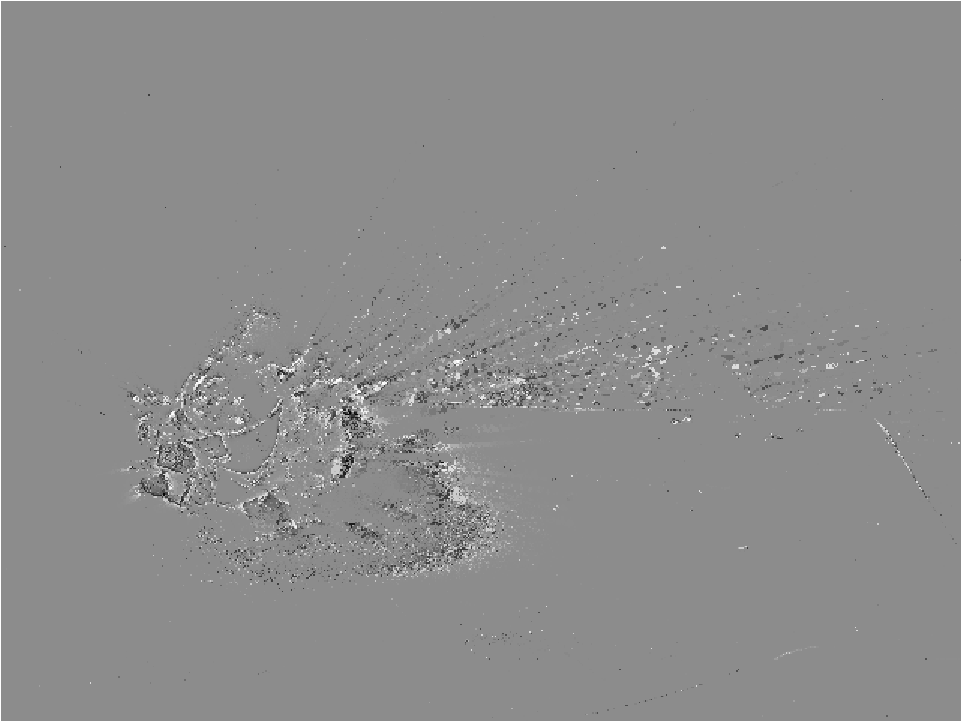}
    
    \caption{Grayscale dynamic sequence of reconstructed images of a cup just before and just after bullet impact from the left. The proposed method not only captures essential edges in the scene, but ballistic dynamics via the gradations in intensities.}
    \label{f:cup}
\end{figure}

\begin{figure}[!h]
    \centering
    \includegraphics[width=.325\textwidth]{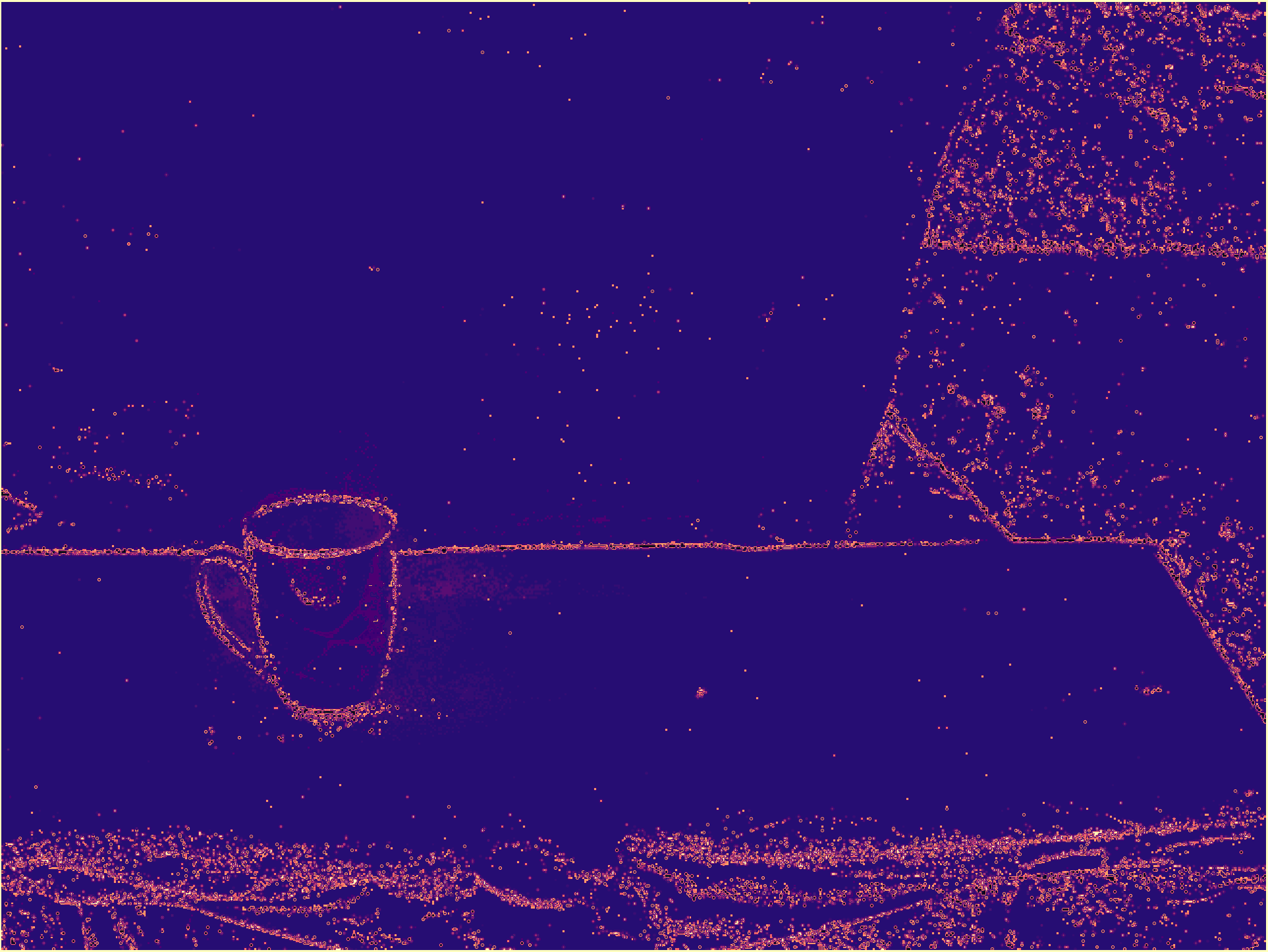}
    \includegraphics[width=.325\textwidth]{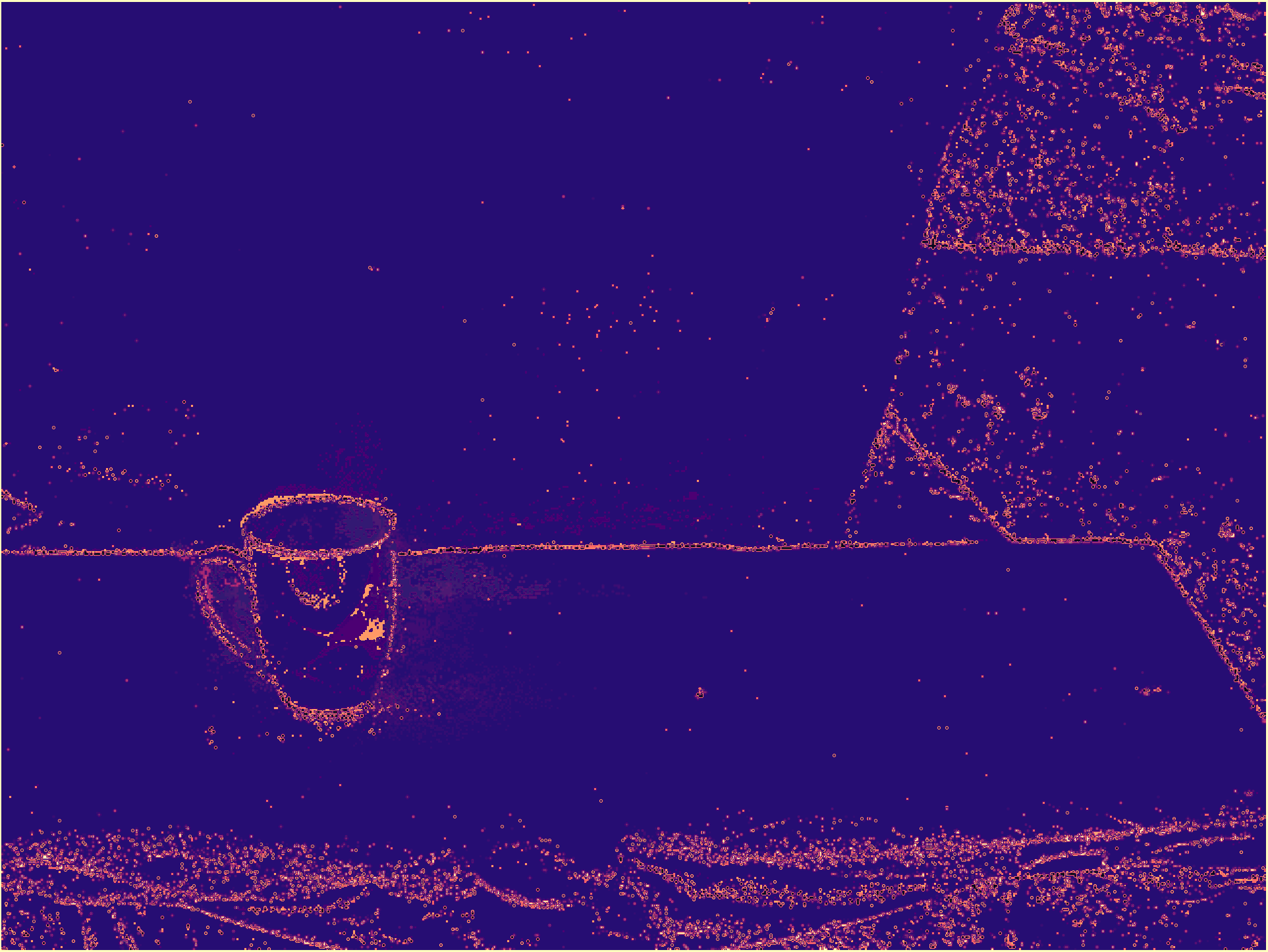}
    \includegraphics[width=.325\textwidth]{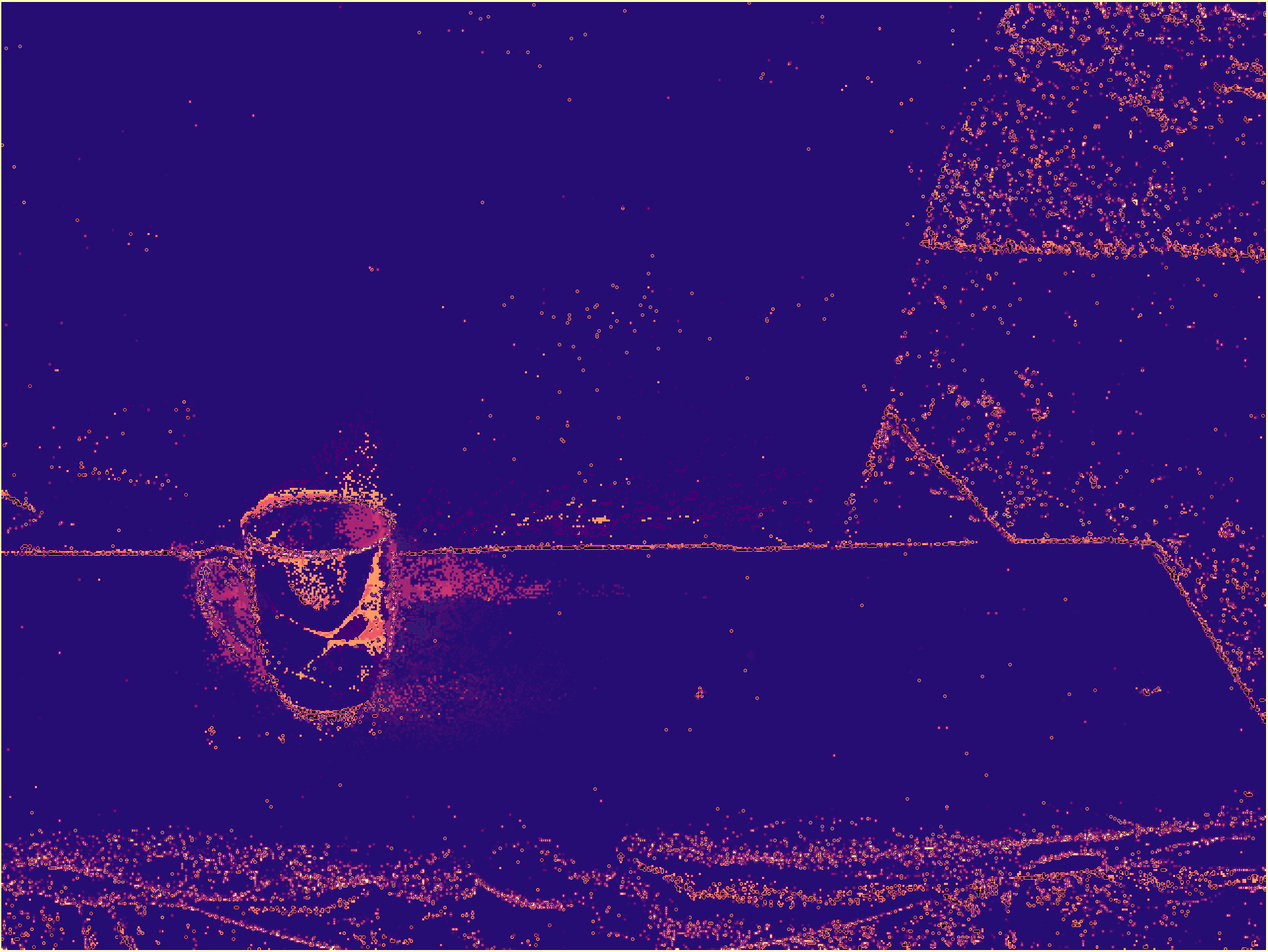}
    \includegraphics[width=.325\textwidth]{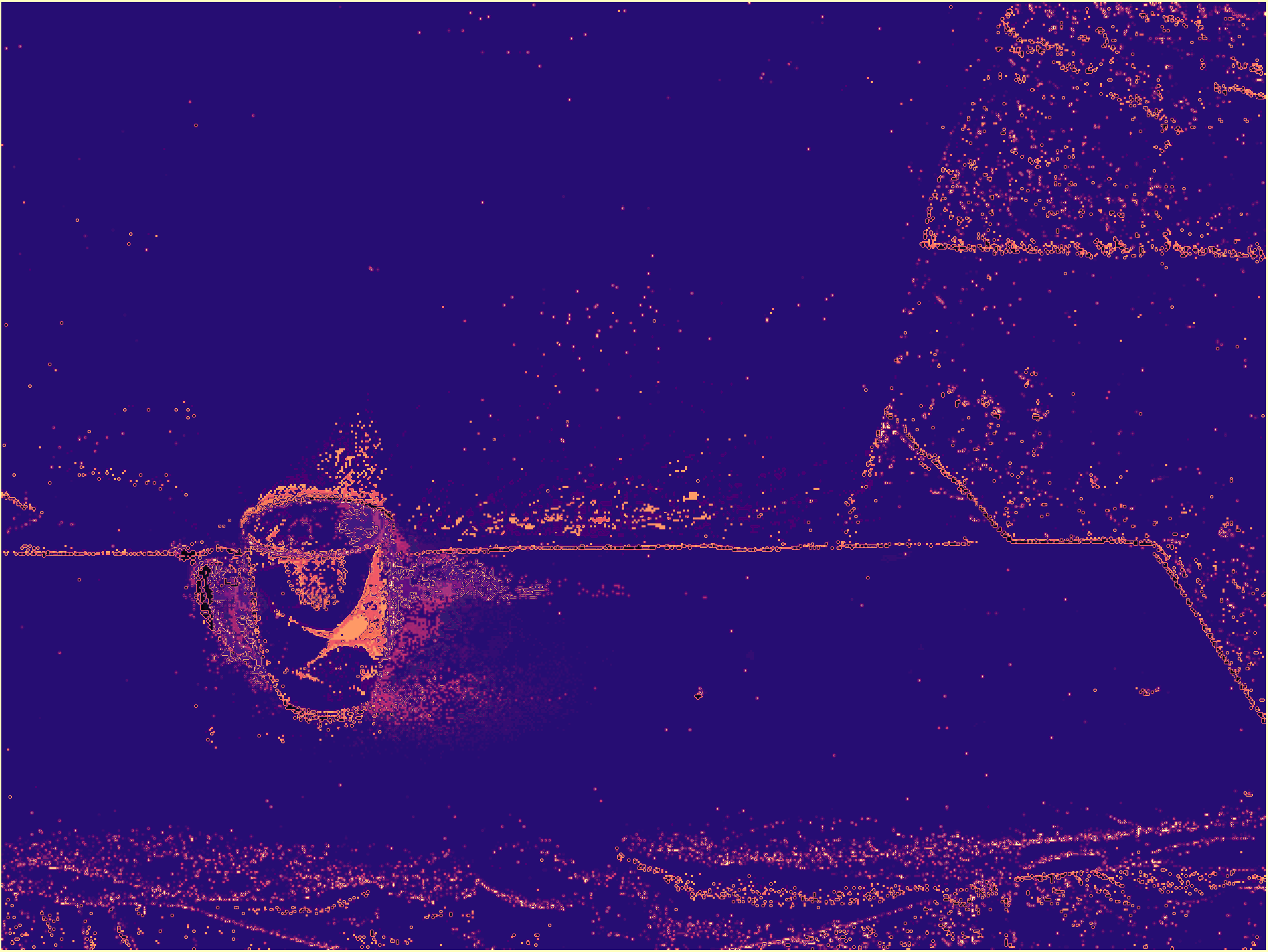}
    \includegraphics[width=.325\textwidth]{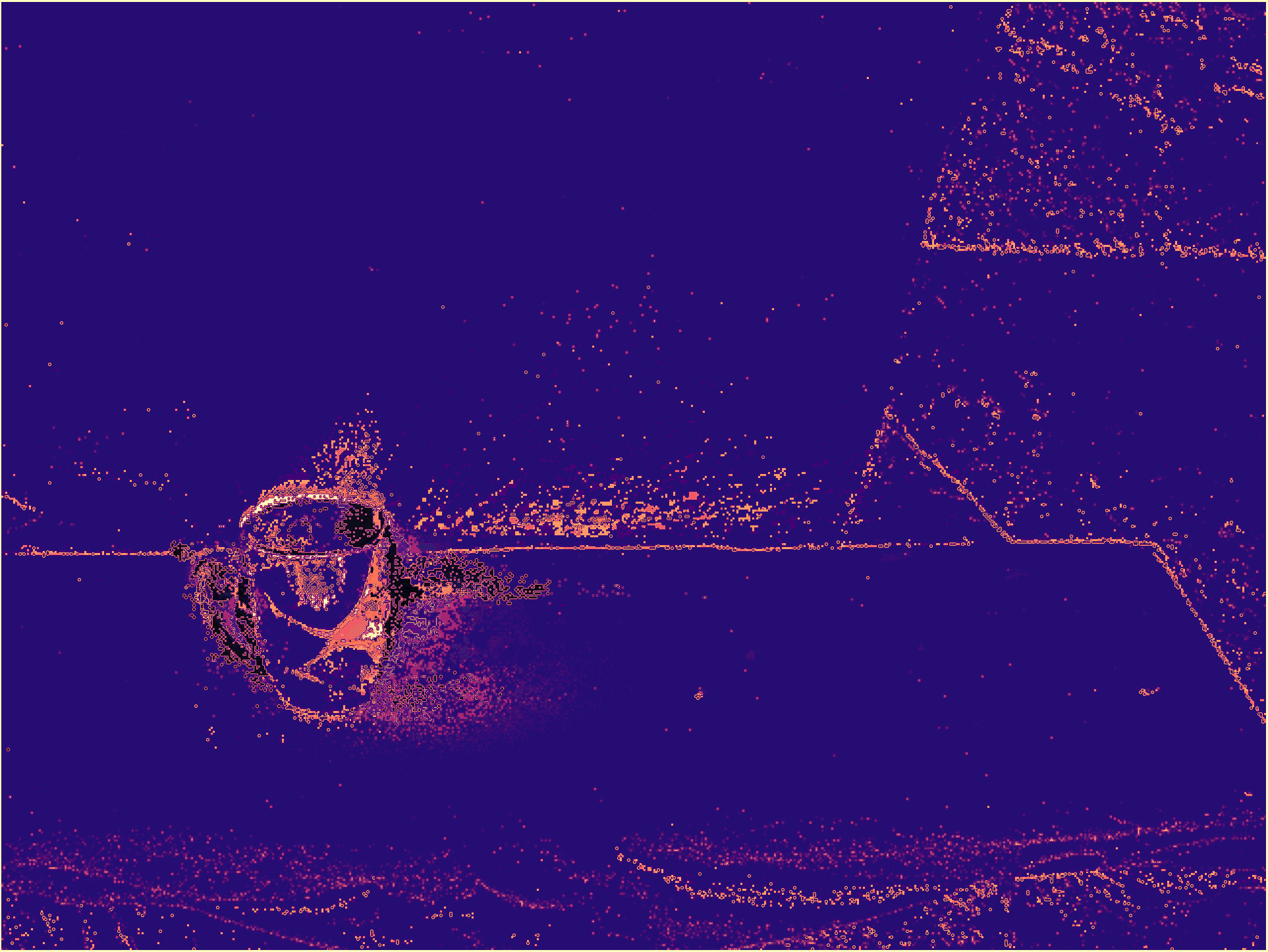} 
    \includegraphics[width=.325\textwidth]{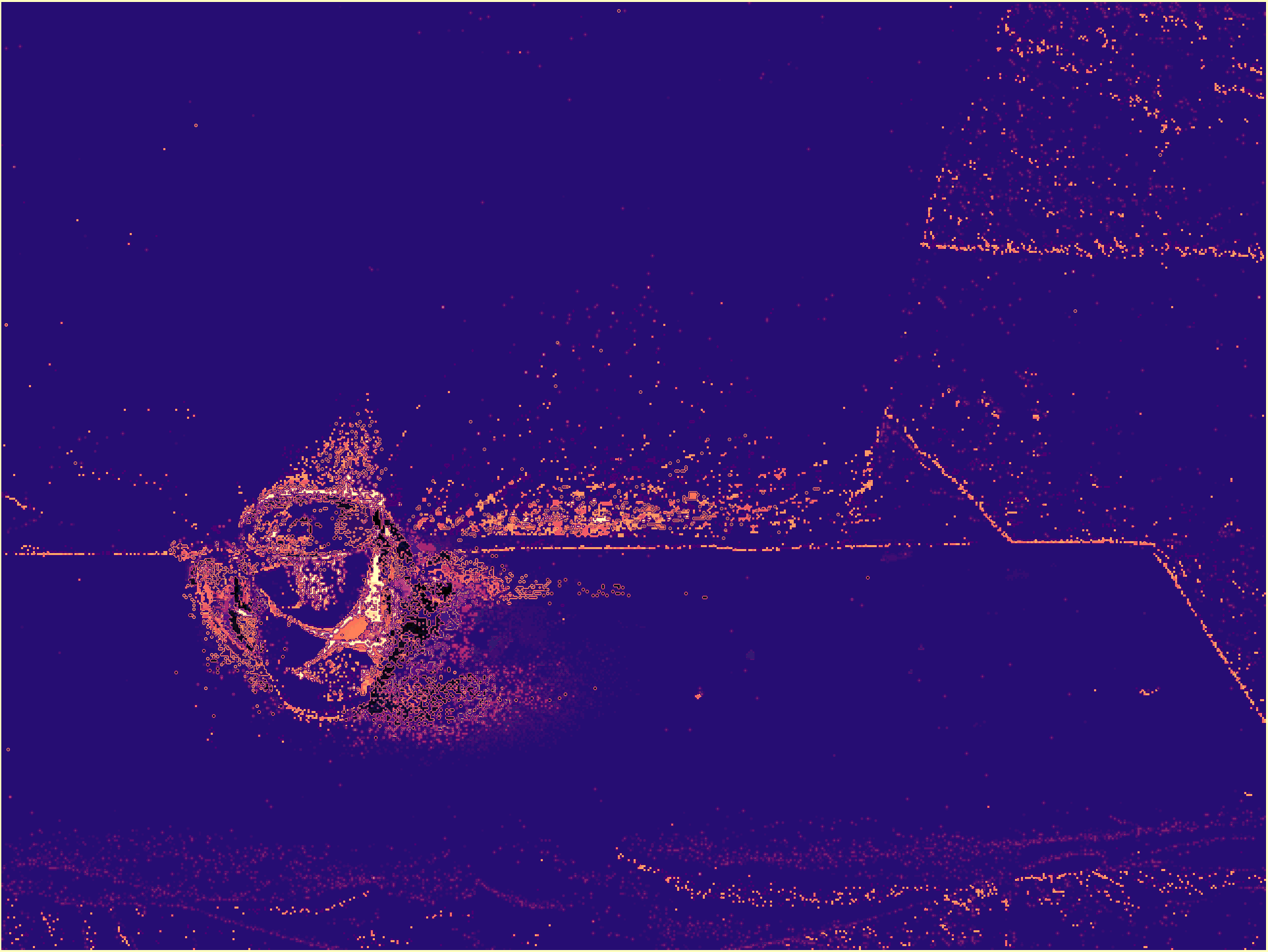}
    \includegraphics[width=.325\textwidth]{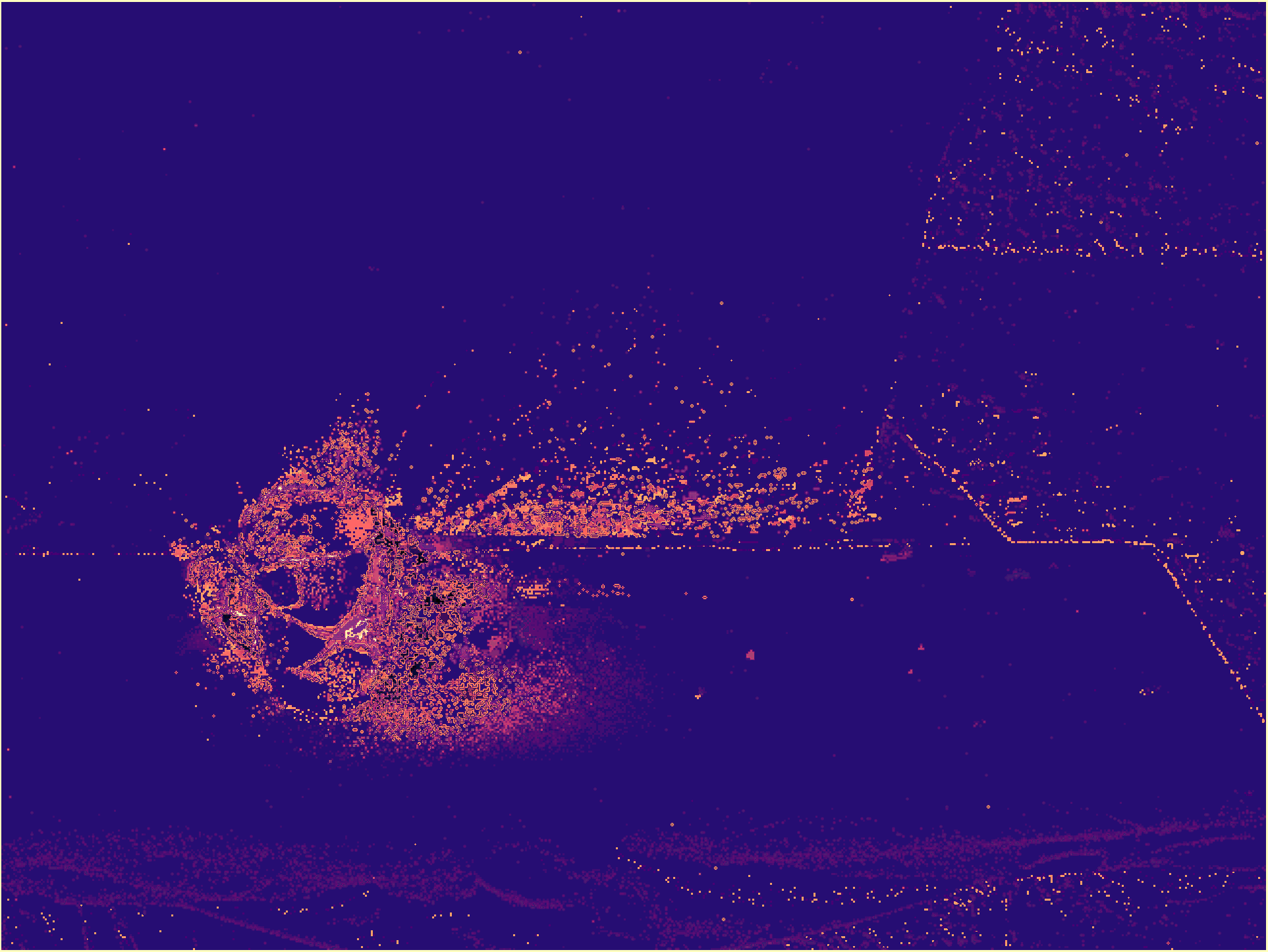}
    \includegraphics[width=.325\textwidth]{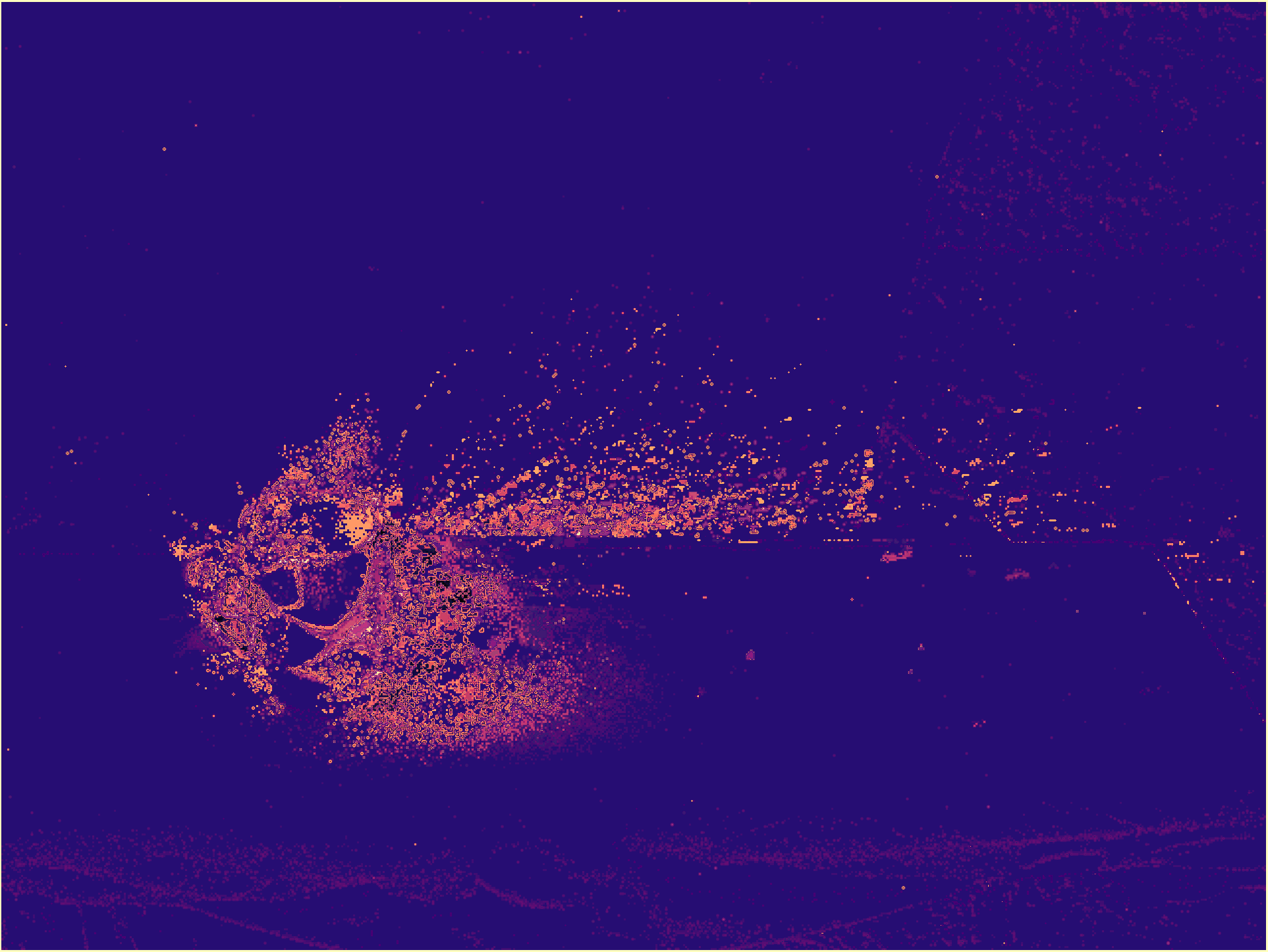}
    \includegraphics[width=.325\textwidth]{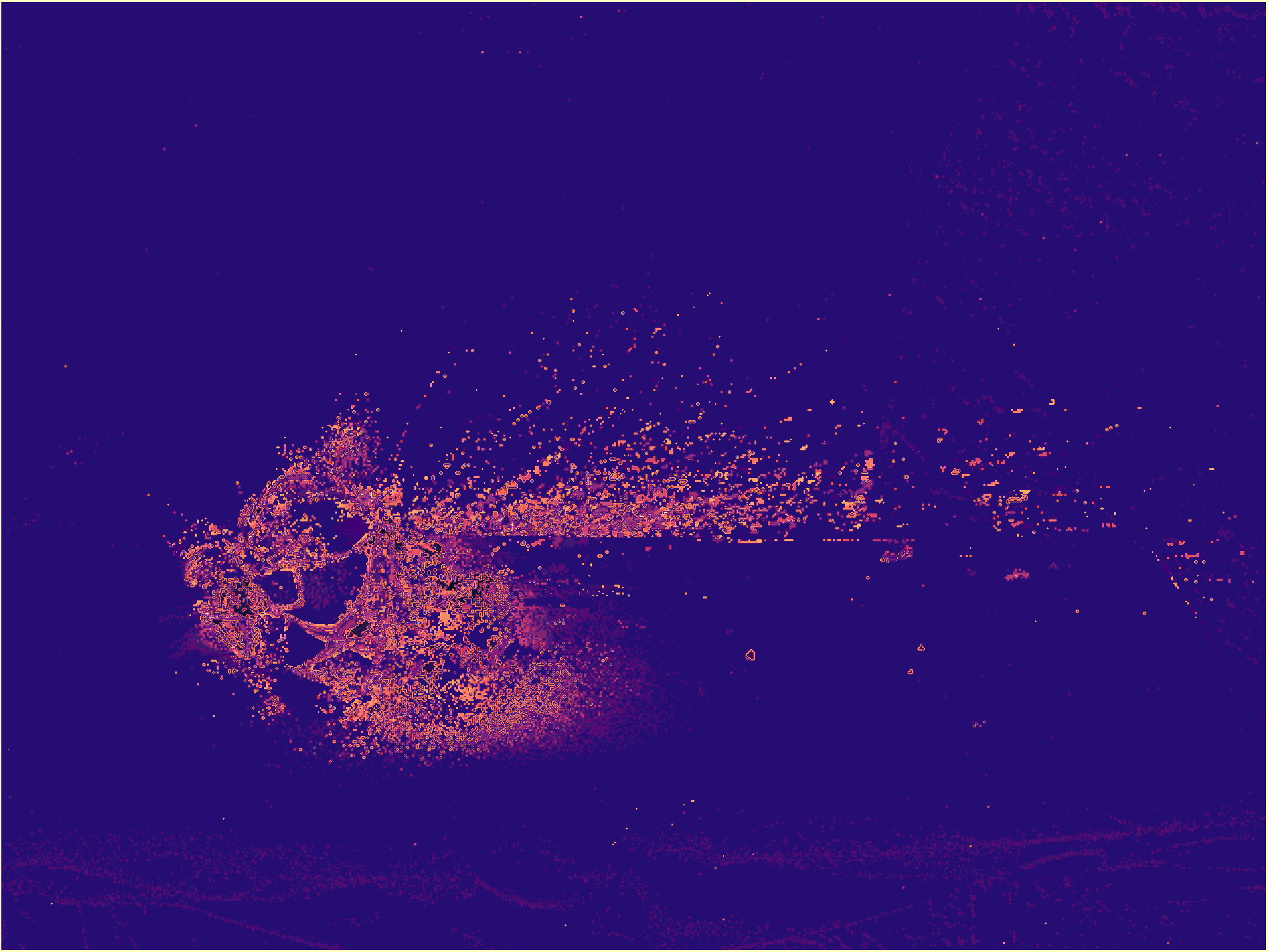}
    \includegraphics[width=.325\textwidth]{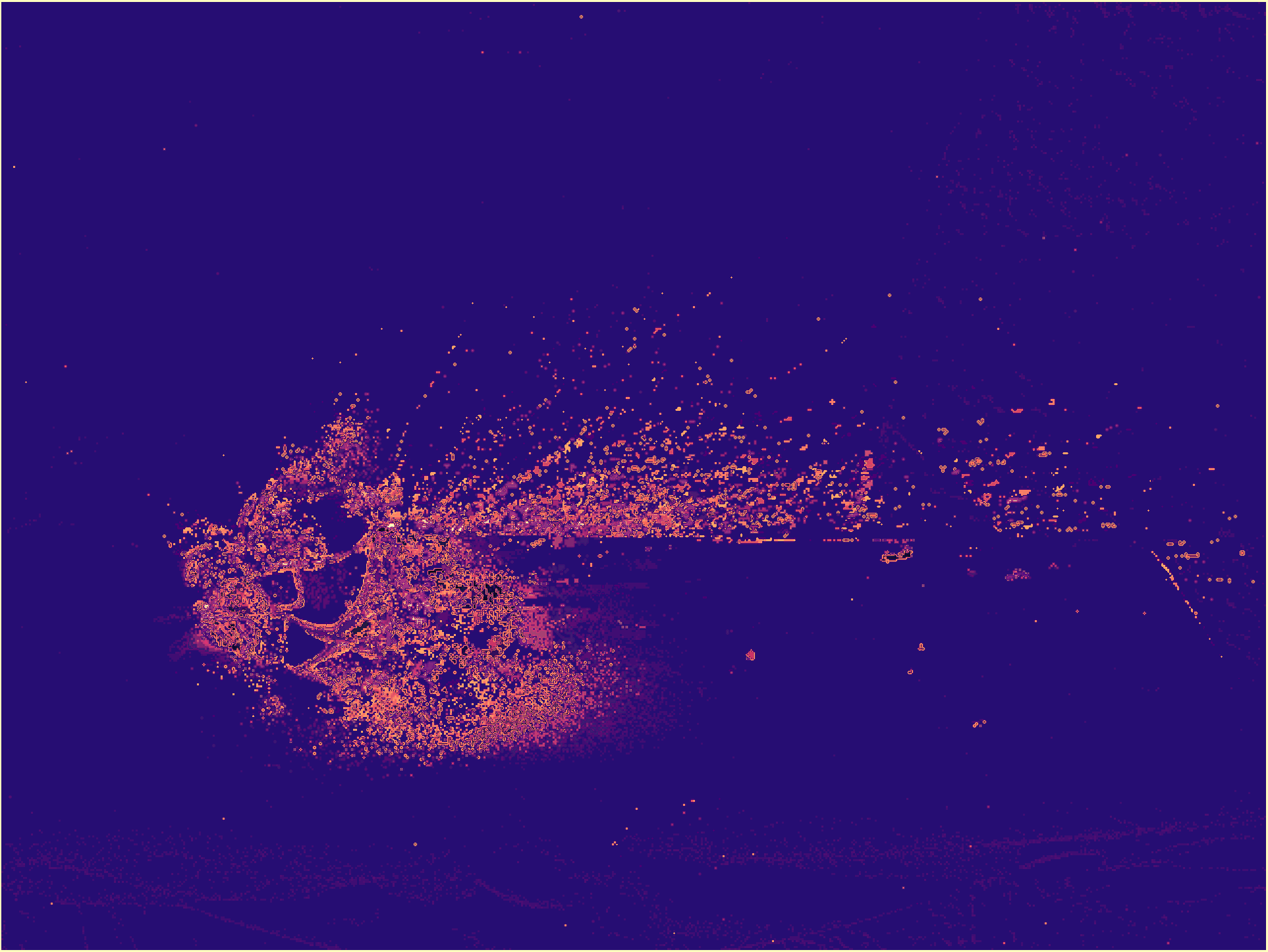}
    \includegraphics[width=.325\textwidth]{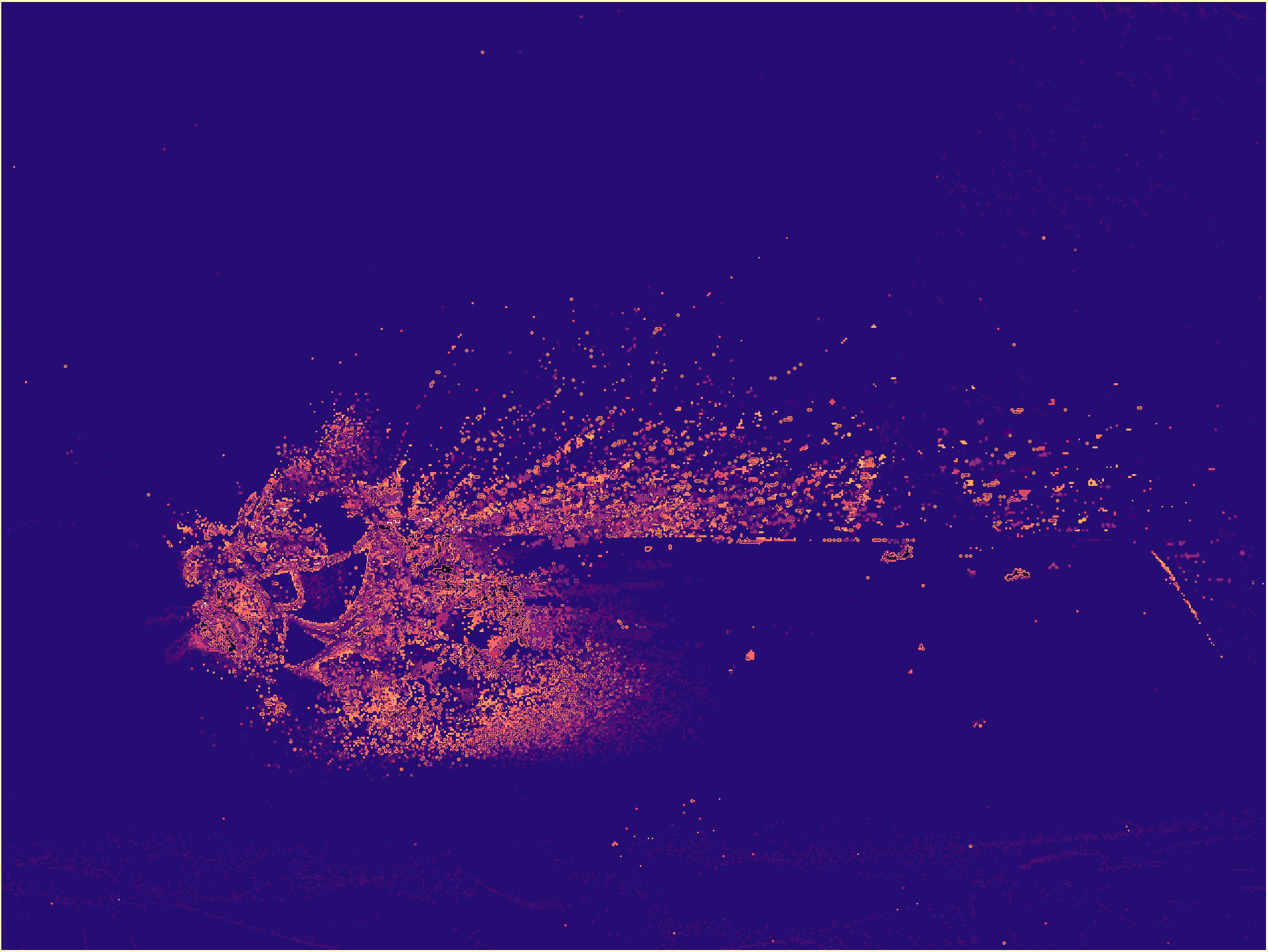} 
    \includegraphics[width=.325\textwidth]{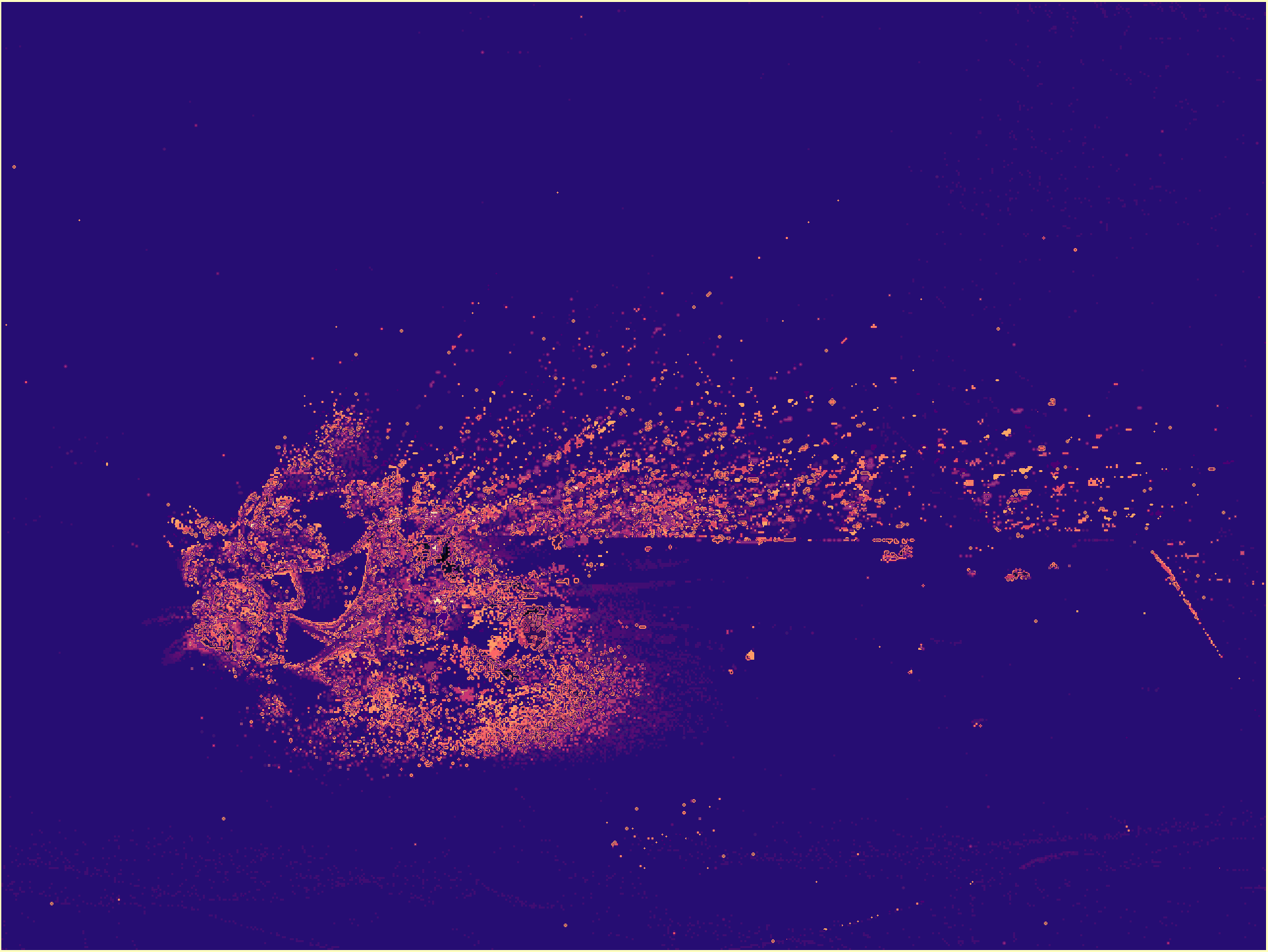}
    \includegraphics[width=.325\textwidth]{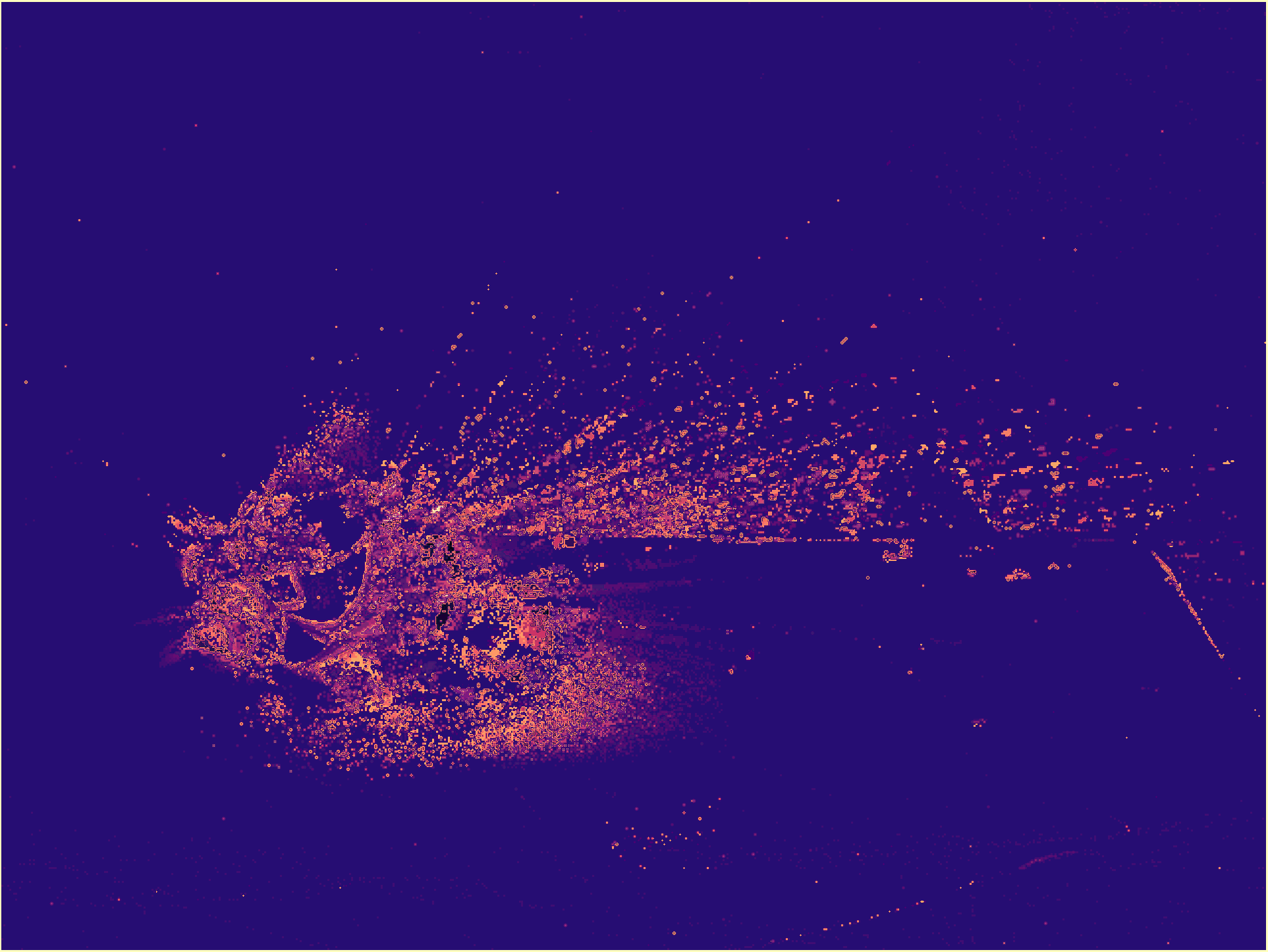}
    \includegraphics[width=.325\textwidth]{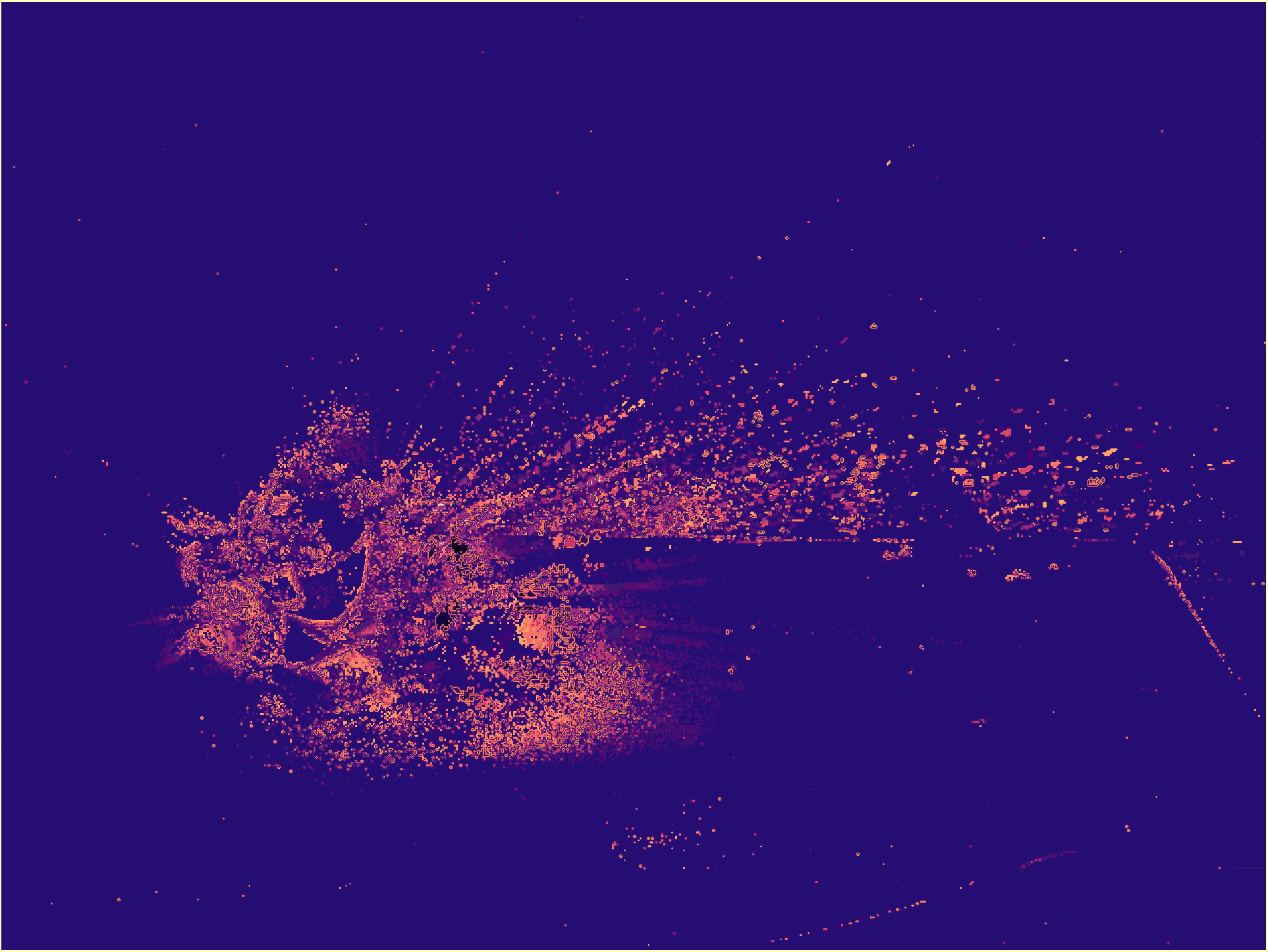} 
    \includegraphics[width=.325\textwidth]{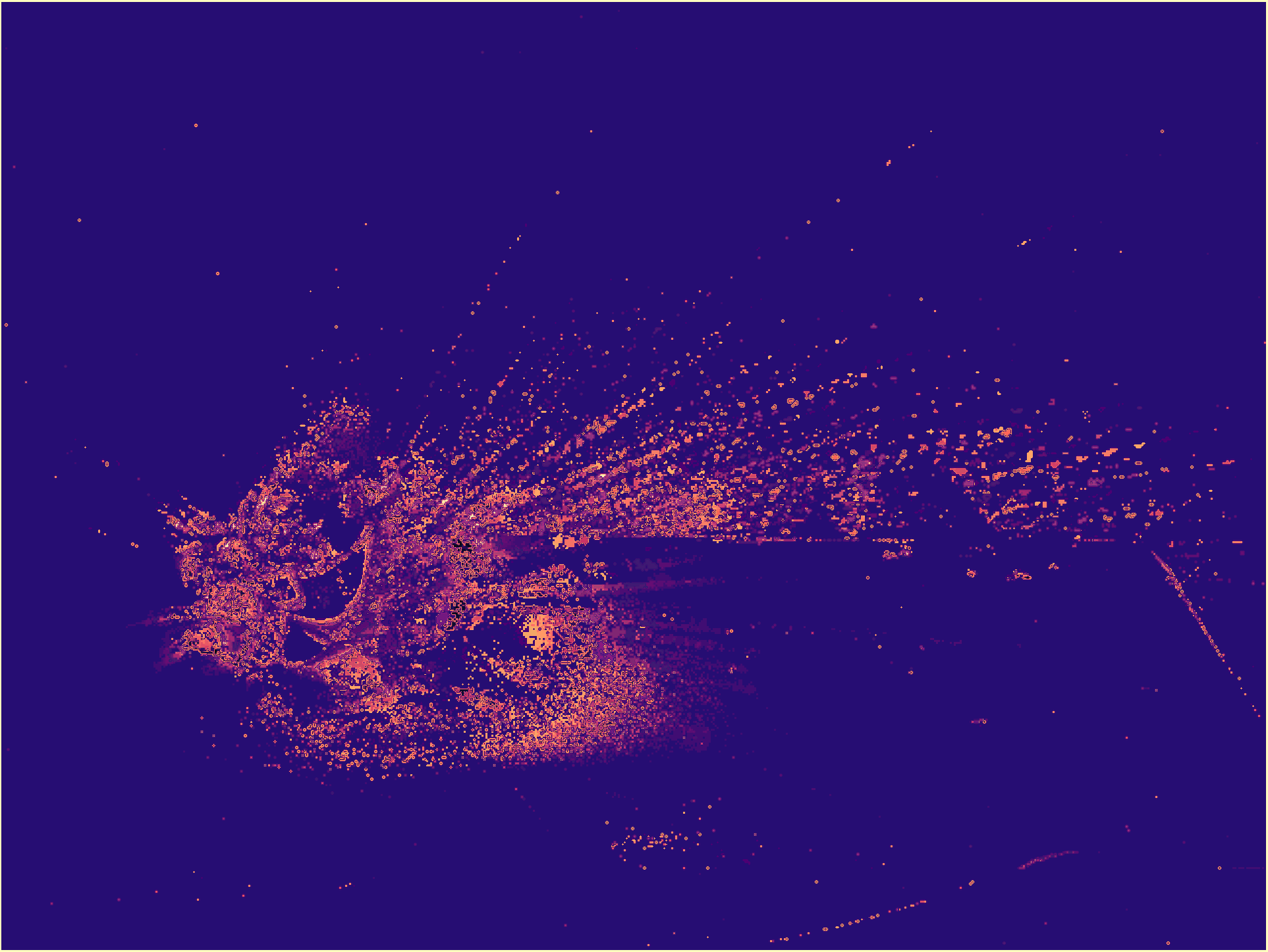}
    \caption{Dynamic ballistic sequence of reconstructed images of cup from Figure \ref{f:cup} with false color map applied to highlight the ballistic dynamics through the cup's initial fracture and shattering.}
    \label{f:cup_cm}
\end{figure}

    \item[Example 3: Falcon Neuro ISS] For our third experiment, we used a dataset from a neuromorhpic camera onboard the International Space Station (ISS) \cite{mcharg2022falcon}. Our image reconstructions for this example used $\nu \approx 2M$ events and the sigmoid function, option (a) above, to compute $\bm{\lambda}$. The Falcon Neuro sensor is composed of two modified commercial DAVIS 240C devices that sit aboard the International Space Station. In Figure \ref{f:issg}, a sequence of overhead images for which we can see our method recovers Honduras' Eastern coastline which runs from the top of center frame down to the bottom of each image. Figure~\ref{f:iss} is the same sequence with a false color map applied to emphasize the rich depth information recovered. Notice that gradation corresponding to depth is visible along the coastline as well as the moving clouds in the upper right quadrant. Not only are high-level topographical and scene features reconstructed, but altitudinal detail for which lighter yellows indicating higher altitudes can be seen on the landmass in the lower left of each image. Figure~\ref{f:fnpi} shows a comparison between our approach and the results in \cite{mcharg2022falcon}.

\begin{figure}[!h]
    \centering
    \subfigure[]{\includegraphics[width=0.325\textwidth]{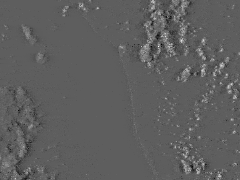}}
    \subfigure[]{\includegraphics[width=0.325\textwidth]{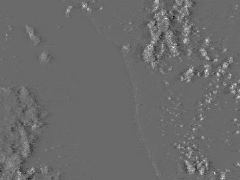}}
    \subfigure[]{\includegraphics[width=0.325\textwidth]{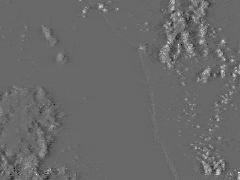}}
    \subfigure[]{\includegraphics[width=0.325\textwidth]{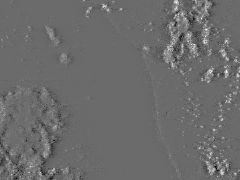}}
    \subfigure[]{\includegraphics[width=0.325\textwidth]{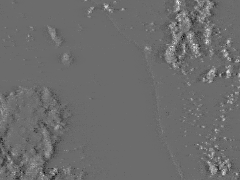}}
    \subfigure[]{\includegraphics[width=0.325\textwidth]{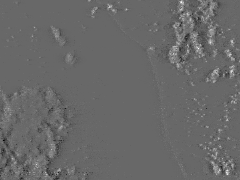}}
    \caption{Grayscale images reconstructed from neuromorphic event-based sensor Falcon Neuro which is aboard the ISS showing the Eastern coastline of Honduras which runs from top to Northwest to Southeast in the right half of each image.} 
    \label{f:issg}
\end{figure}

\begin{figure}[!h]
    \centering
    \subfigure[]{\includegraphics[width=0.325\textwidth]{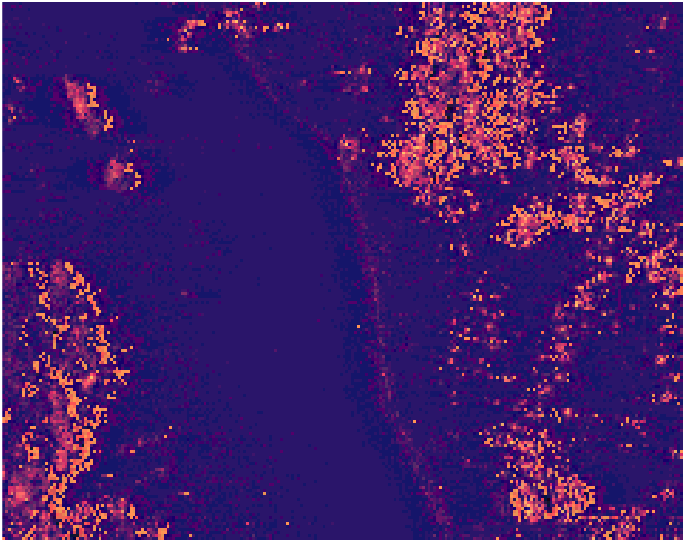}}
    \subfigure[]{\includegraphics[width=0.325\textwidth]{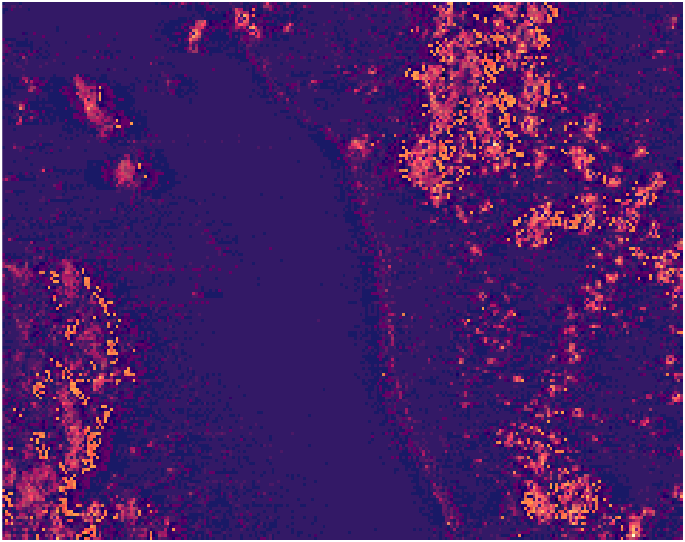}}
    \subfigure[]{\includegraphics[width=0.325\textwidth]{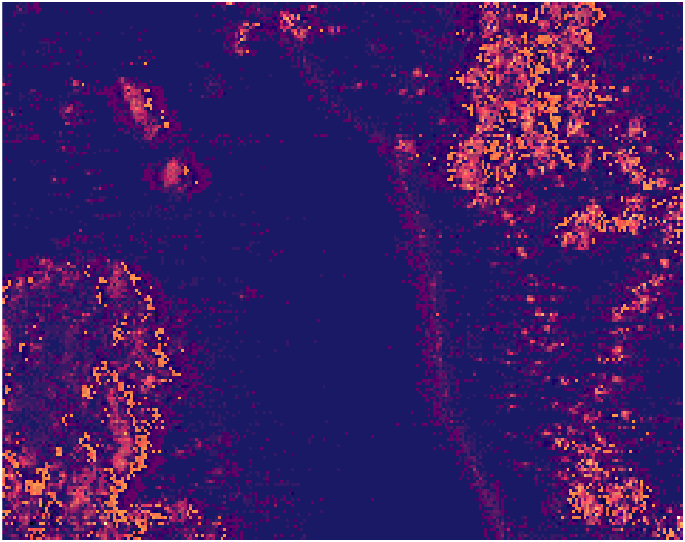}}
    \subfigure[]{\includegraphics[width=0.325\textwidth]{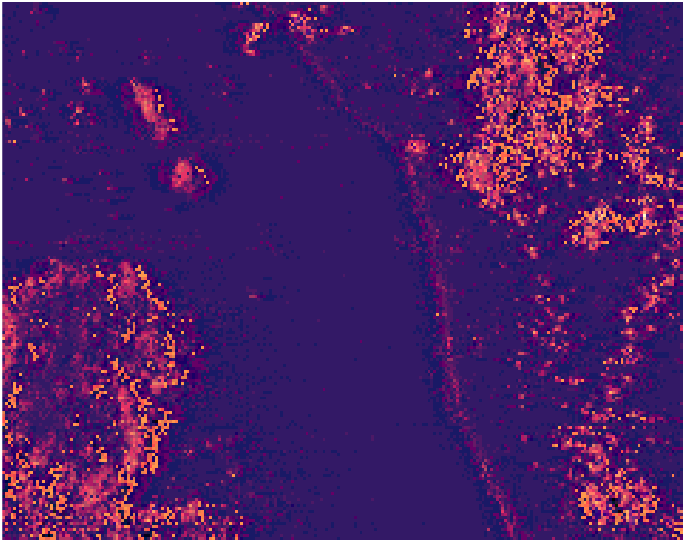}}
    \subfigure[]{\includegraphics[width=0.325\textwidth]{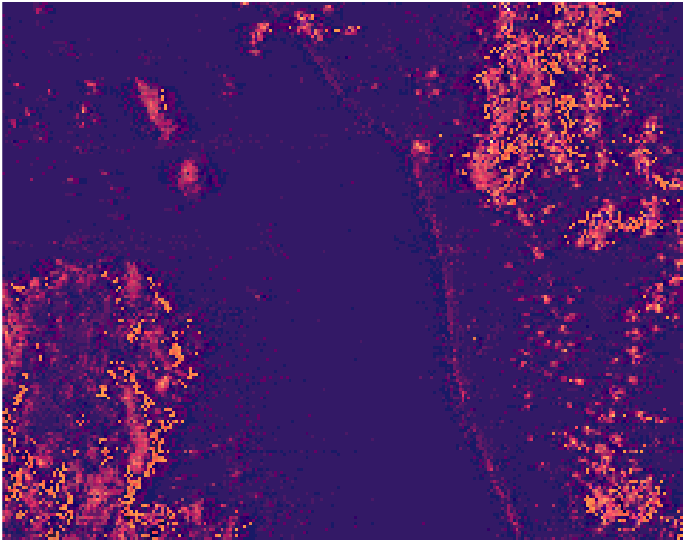}}
    \subfigure[]{\includegraphics[width=0.325\textwidth]{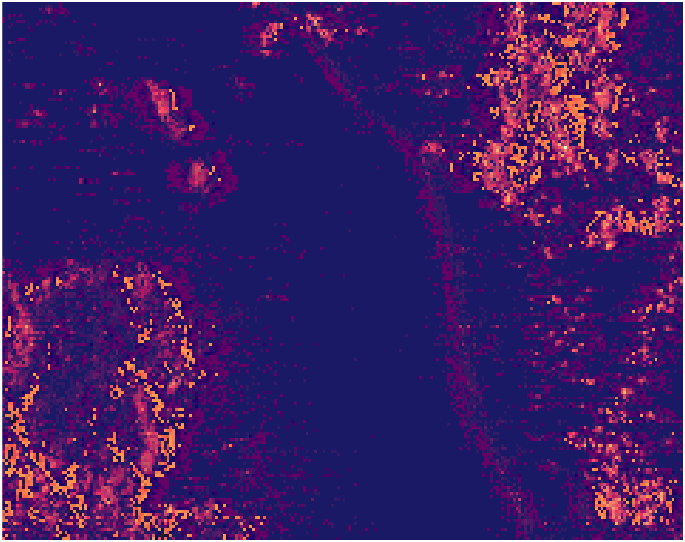}}
    \caption{Sequence of reconstructed overhead images from Figure \ref{f:issg} with false color map applied to emphasize reconstruction details.}
    \label{f:iss}
\end{figure}

\begin{figure}[!h]
    \centering
    \subfigure[]{\includegraphics[width=0.325\textwidth]{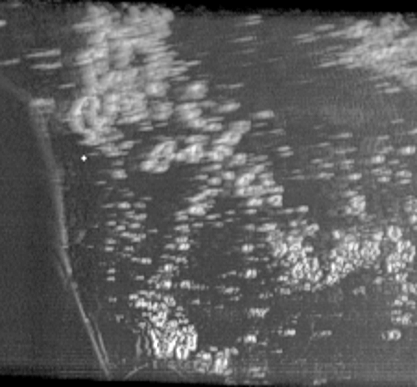}}
    \subfigure[]{\includegraphics[width=0.325\textwidth]{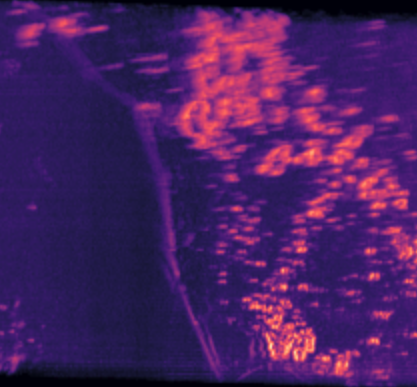}}
    \caption{Single image from original reference images presented in \cite{mcharg2022falcon} included here to demonstrate the efficacy our method in Figures \ref{f:issg} and \ref{f:iss}.}
    \label{f:fnpi}
\end{figure}

\end{description}

\section*{Conclusion} \label{s:con}
In conclusion, this study introduces a pioneering approach to neuromorphic image reconstruction, incorporating temporal information for individual pixel intensity reconstruction through an optimization problem, yielding unique solutions governed by the regularization parameter $\bm \lambda$.
\begin{enumerate}
    \item The pixel-by-pixel optimization inherent in our method raises scalability concerns, particularly in higher camera resolutions where dynamics are prevalent across the majority of pixels.
    \item Despite providing a systematic approach for selecting regularization parameters for each pixel/time step, we cannot claim optimality in these parameter choices.
\end{enumerate}
To mitigate scalability challenges, we propose a strategy involving the exclusion of pixels with significantly lower event counts. Furthermore, we envision the development of a parameter learning framework aimed at optimizing regularization parameters to enhance overall performance.

Follow on to this work may include combining the method presented in this paper with the spatial method proposed in \cite{zhang2021formulating}.


\bibliographystyle{plain}
\bibliography{refs.bib}

\end{document}